\pdfoutput=1
\documentclass{article}

\usepackage{iclr2027_conference,times}

\usepackage{graphicx}
\usepackage{wrapfig}
\usepackage{needspace}
\usepackage{amsmath}
\usepackage{amsfonts}
\usepackage{amssymb}
\usepackage{bm}
\usepackage{algorithm}
\usepackage{algorithmic}
\usepackage[inline]{enumitem}
\usepackage{float}
\usepackage{hyperref}
\definecolor{linkpurple}{HTML}{8C6BB1}
\hypersetup{colorlinks=true, linkcolor=linkpurple, citecolor=linkpurple, urlcolor=linkpurple}
\definecolor{tokRW}{HTML}{985841}%
\definecolor{tokLW}{HTML}{6D6A42}%
\definecolor{tokRE}{HTML}{AA4D27}%
\definecolor{tokLE}{HTML}{6F6A35}%
\definecolor{tokTO}{HTML}{3F6E8B}%
\definecolor{tokLO}{HTML}{485A6C}%
\definecolor{tokPU}{HTML}{4B3B8C}%
\definecolor{tokPP}{HTML}{925A55}%
\definecolor{tokIM}{HTML}{A1514B}%
\DeclareRobustCommand{\modcolor}[2]{\textcolor{#1}{#2}}
\usepackage{url}
\usepackage{xspace}
\usepackage{comment}
\usepackage{booktabs}
\usepackage{tabularx}
\usepackage{array}
\usepackage{etoolbox}
\AtBeginEnvironment{table}{\setlength{\abovecaptionskip}{0pt}\setlength{\belowcaptionskip}{10pt}}
\AtBeginEnvironment{table*}{\setlength{\abovecaptionskip}{0pt}\setlength{\belowcaptionskip}{10pt}}
\newcommand{\method}{EgoX\xspace}

\newcommand{\parab}[1]{\noindent\textbf{#1}}

\let\cite\citep

\title{EgoX: Egocentric Cross-Embodiment\\Manipulation with Embodiment Dreaming}

\author{%
Yaru Niu$^{1,*}$, Binghong Chen$^{1,*}$, Changwei Yao$^{1,*}$, Shangtao Li$^{1,*}$, Zhenlong Fang$^{4}$ \\
\bfseries Revanth Krishna Senthilkumaran$^{1,2}$, Shuai Zhou$^{1}$, Yuemin Mao$^{1}$ \\
\bfseries Kalie Ching$^{1,2}$, Po-Ju Chao$^{1}$, Yudong Luo$^{1}$, Peiqi Yu$^{1}$, Heng Zhang$^{1}$ \\
\bfseries Hao Zhang$^{1,3}$, Bingqing Chen$^{2}$, Chen Qiu$^{2}$, H. Eric Tseng$^{3}$ \\
\bfseries Ye Yuan$^{5}$, Umar Iqbal$^{5}$, Changliu Liu$^{1}$, Jonathan Francis$^{1,2}$, Ding Zhao$^{1}$ \\[4pt]
\small $^{1}$\,Carnegie Mellon University \quad $^{2}$\,Bosch Center for Artificial Intelligence \\
\small $^{3}$\,The University of Texas at Arlington \quad $^{4}$\,University of California, Berkeley \quad $^{5}$\,NVIDIA \\[2pt]
\small $^{*}$\,Equal contribution \\[2pt]
\small\href{https://egox-dream.github.io}{https://egox-dream.github.io}}

\iclrfinalcopy
\makeatletter
\AtBeginDocument{\global\let\lhead\@gobble}
\makeatother
\renewcommand{\headrulewidth}{0pt}

\begin{document}

\begingroup
\setlength{\tabcolsep}{0pt}
\maketitle
\endgroup
\vspace{-6pt}

\begin{figure}[H]
    \centering
    \includegraphics[width=\linewidth]{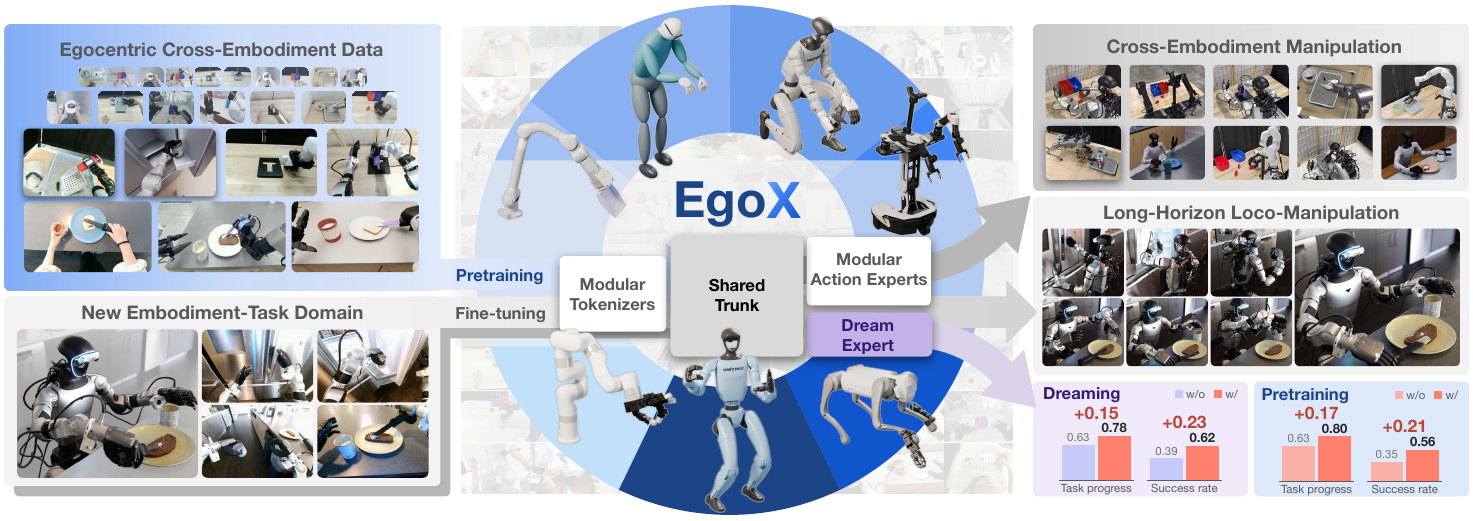}
    \caption{\small
    \textbf{Overview of EgoX.} EgoX pretrains on egocentric demonstrations from humans, physical robots,
    and simulated robots, then fine-tunes on a target embodiment--task domain (left).
    MED connects embodiment-specific modular tokenizers and action experts through a shared Transformer
    trunk, while a training-only dream expert predicts future embodiment latents (center). Policies perform
    manipulation across diverse bodies and end-effectors, including long-horizon humanoid
    loco-manipulation (right, top and middle). Cross-embodiment pretraining and embodiment dreaming
    improve average task progress and success rate (right, bottom).
    }
    \label{fig:framework}
\end{figure}

\begin{abstract}
    
Generalist manipulation policies can benefit from the complementary experience of humans, real robots, and simulated agents. However, differences in morphology, sensing, and control interfaces make it difficult to learn effectively from their combined demonstrations. We introduce EgoX, an egocentric cross-embodiment learning framework that shares manipulation structure through fine-grained, semantically corresponding observation and action modalities while keeping sensing and control interfaces embodiment-specific. EgoX pretrains a Modular Embodiment Dreamer (MED) on this mixture and fine-tunes it on target-embodiment demonstrations. Within each embodiment, MED assigns a dedicated tokenizer to each observation modality and a dedicated expert to each action modality, with a Transformer trunk shared across embodiments. A training-only embodiment-dreaming objective predicts latent representations of future visual and proprioceptive states, encouraging the shared trunk to model each embodiment's dynamics. Across 13 task--embodiment pairs spanning five tasks and six robot embodiments, including five physical robots and one simulated humanoid, cross-embodiment pretraining raises MED's average normalized task progress from 0.63 to 0.80 over training from scratch. The gains are particularly pronounced under out-of-distribution conditions, where average task progress increases from 0.53 to 0.77. Ablations demonstrate the benefits of broader pretraining mixtures, modular interfaces, and embodiment dreaming, with dreaming producing more similar latent geometry across embodiments.

\end{abstract}

\section{Introduction}
\label{sec:introduction}

Large-scale pretraining has driven rapid progress in natural language processing~\cite{devlin2019bert,brown2020language} and computer vision~\cite{radford2021learning,dosovitskiy2021image}, demonstrating the potential of learning reusable knowledge from diverse data. A corresponding ambition in robotics is to develop generalist models that accumulate manipulation knowledge across tasks and environments and can be adapted to different robot embodiments. We envision a future populated by robots with complementary physical capabilities: tabletop arms, wheeled mobile manipulators, humanoids, and quadrupeds offer different combinations of mobility, reach, and workspace access, while grippers, dexterous hands, and specialized tools support different forms of object interaction~\cite{pfeifer2007self,baines2024robots,kirschner2025categorizing}. Their different ways of perceiving and interacting with the world also provide complementary experience for learning. However, collecting data and training models separately for every embodiment, task, and scene is costly and limits the reuse of this experience. How can we draw on experience across these embodiments to develop shared manipulation knowledge while allowing each robot to adapt it to its own body and environment?

Several complementary data sources offer a path toward this goal. Cross-embodiment robot learning draws on demonstrations collected across different bodies, sensors, and action spaces~\cite{o2024open,team2024octo,kim2024openvla,doshi2024scaling,wang2024scaling,team2025gemini15}. Egocentric human data captures diverse physical interactions in real-world settings beyond existing robot deployment environments~\cite{kareer2025egomimic,niu2025human2locoman,qiu2025humanoid,zheng2026egoscale}. Simulation enables scalable data collection and has supported policy pretraining, sim-to-real transfer, and co-training with real-world demonstrations~\cite{deshpande2026molmobot,bjorck2025gr00t,maddukuri2025sim,lin2026humanoidmimicgen}. Together, these sources can broaden the objects, scenes, and interaction strategies available for learning beyond the demonstrations collected for any individual robot. Realizing this potential, however, requires training recipes that can make effective use of heterogeneous experience. Demonstrations from different embodiments encode interactions through distinct morphologies, viewpoints, proprioception, action semantics, and motion dynamics, making it challenging to turn their combined experience into useful manipulation capabilities across robots.

\begin{figure}[!t]
    \centering
    \includegraphics[width=\linewidth]{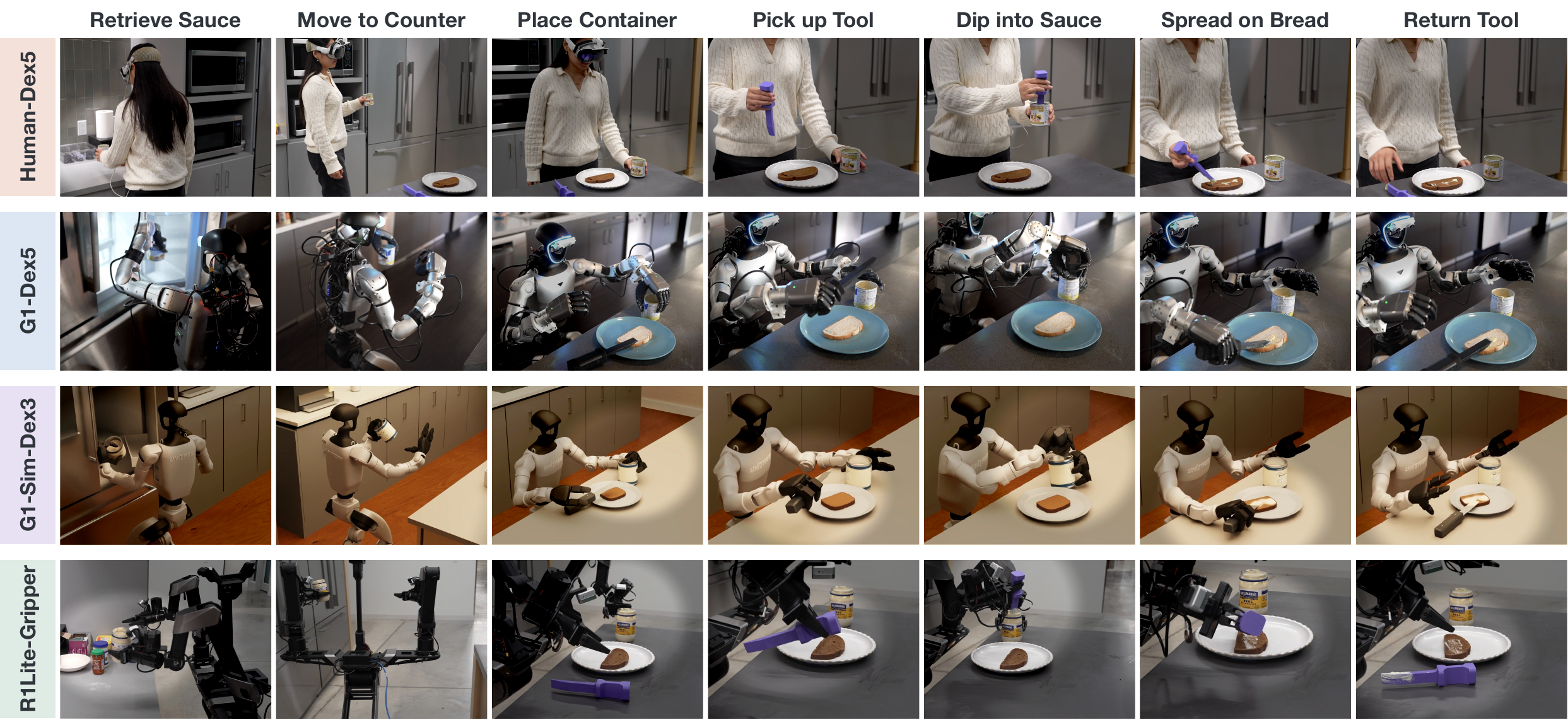}
    \caption{\small
    \textbf{Shared task structure across heterogeneous embodiments.}
    Rows show a \texttt{Human-Dex5} demonstration and rollouts of robot policies trained
    with EgoX for \texttt{G1-Dex5}, \texttt{G1-Sim-Dex3}, and \texttt{R1Lite-Gripper}
    on \texttt{Kitchen-Bread}. Columns show corresponding task stages.
    Soft spotlights highlight the active interaction regions.
    }
    \label{fig:kitchen_bread_cross_embodiment}
\end{figure}

Recent work has examined scaling laws and data composition for robot manipulation~\cite{lin2025data,lin2026systematic,trilbm2025careful}. Alongside these questions, we focus on how experience from heterogeneous embodiments should be collected, represented, and used for shared learning. Our central concern is the design of representations and learning interfaces that allow diverse demonstrations to contribute to a common manipulation model while preserving the differences that matter for each embodiment's sensing and control. We treat humans, real robots, and simulated agents as embodiments on equal footing, each contributing its own observations and actions. Through joint pretraining and adaptation to target embodiments, we investigate how this shared experience can improve downstream manipulation across different robot bodies and end-effectors. Fig.~\ref{fig:kitchen_bread_cross_embodiment} illustrates this perspective through corresponding stages of \texttt{Kitchen-Bread} performed by human, real-robot, and simulated embodiments.

We introduce EgoX (Fig.~\ref{fig:framework}), an egocentric cross-embodiment learning framework built around a single methodological premise: transfer should occur through semantically corresponding modalities, while sensing and control interfaces remain embodiment-specific. The framework has three components. First, \method aligns human, real-robot, and simulated demonstrations in a unified observation-action representation, grouping heterogeneous signals into anatomically or functionally relevant modalities and expressing poses in a unified egocentric frame system.
Second, MED (Modular Embodiment Dreamer) adopts a fine-grained modular design: within each embodiment, each observation modality has its own tokenizer and each action modality its own expert, while a shared Transformer trunk integrates information across modalities and embodiments.
Third, MED uses \textit{embodiment dreaming}, in which an embodiment tokenizer encodes current egocentric images and embodiment-specific proprioception, while its exponential-moving-average (EMA) copy encodes the corresponding future observations into embodiment latents. A dream expert predicts these latents from the current shared representation. This training-only expert encourages the trunk to model each embodiment's dynamics, without pixel reconstruction or deployment-time computation.

We evaluate EgoX on five manipulation tasks, from long-horizon loco-manipulation to precise insertion and tool use (\texttt{Kitchen-Bread}, \texttt{Kitchen-Cleanup}, \texttt{Sort}, \texttt{Insert}, and \texttt{Scoop}), across six robot embodiments: \texttt{G1-Dex5}, \texttt{xArm7-Dex4}, \texttt{G1-Sim-Dex3}, \texttt{LocoMan-Dex2}, \texttt{R1Lite-Gripper}, and \texttt{Z1-Tool}. Human demonstrations provide an additional pretraining source. Our experiments compare MED variants with baseline policies across all tasks and ablate the pretraining mixture, embodiment dreaming objective, and MED's modular design. Cross-embodiment pretraining raises MED's average task progress from 0.63 to 0.80, yielding the highest average score among the evaluated methods and the best performance on 12 of 13 task--embodiment pairs. The gains from pretraining are particularly pronounced under out-of-distribution (OOD) conditions, where average task progress increases from 0.53 to 0.77. Broader pretraining mixtures transfer better, and embodiment dreaming raises task progress from 0.63 to 0.78 across seven settings while making the trunk's latent geometry more similar across embodiments.
In summary, our contributions are:

\begin{enumerate}
    \item \textbf{A unified, modality-structured representation for egocentric cross-embodiment manipulation.} \method unifies human, real-robot, and simulated demonstrations into a common observation--action structure based on anatomically or functionally corresponding modalities, while retaining embodiment-specific sensing and control interfaces.
    \item \textbf{MED, a modular cross-embodiment Transformer with embodiment dreaming.} Within each embodiment, MED assigns dedicated tokenizers and action experts to individual observation and action modalities, connected through a shared Transformer trunk. Embodiment dreaming encourages the trunk to model each embodiment's transition dynamics through future-latent prediction used only during training.
    \item \textbf{A systematic study of cross-embodiment transfer.} Experiments across five tasks and six diverse robot embodiments demonstrate the benefits of cross-embodiment pretraining, particularly under OOD conditions. Ablations show benefits from broader pretraining mixtures, modular interfaces, and embodiment dreaming, with dreaming also producing more similar latent geometry across embodiments.
\end{enumerate}

\section{Related Work}
\label{sec:related_work}
\parab{Cross-Embodiment Learning for Robotic Manipulation.}
Large robot datasets and generalist policies demonstrate the value of learning across tasks and robot embodiments~\cite{o2024open,kim2024openvla,black2024pi_0,intelligence2025pi_,team2025gemini}, with recent work incorporating human data~\cite{luo2026beingh05,bjorck2025gr00t}.
Prior methods address embodiment heterogeneity through shared-trunk policies with embodiment-specific tokenizers or action heads~\cite{doshi2024scaling,niu2025human2locoman,wang2024scaling}, a single embodiment-conditioned action head~\cite{li2026xdiffvla}, or image editing that aligns robot appearance~\cite{chen2024mirage,lepert2024shadow}.
Some methods cover a narrow range of robot morphologies and end-effectors, as in Human2LocoMan~\cite{niu2025human2locoman}, which studies human-to-robot transfer on a quadrupedal platform in two manipulation modes, without considering locomotion or loco-manipulation.
Other approaches adopt coarser observation and action interfaces, leaving correspondences between anatomical or functional modalities less explicit.
To address this, EgoX systematically decomposes observations and actions into anatomically or functionally corresponding modalities with modality-specific reference frames to support highly diverse bodies and end-effectors.
This includes root-frame velocity commands for loco-manipulation and wrist-relative keypoint positions for hands, grippers, and tools.
MED implements this formulation through fine-grained modality tokenizers and action experts around a shared trunk and adds embodiment dreaming for future latent prediction.

\parab{Robot Learning from Egocentric Data.}
Egocentric human data offers a scalable source of manipulation experience spanning diverse tasks, objects, and environments. Recent advances in wearable devices, such as XR headsets and smart glasses, have improved the quality and scale of such data~\cite{chen_arcap_nodate,hoque2025egodex,punamiya2026egoverse}.
Human videos have been used to learn transferable priors through visual, affordance, and latent-action objectives~\cite{nair2022r3m,srirama2024hrp,ye2024latent,bu_univla_2025,zhang_clap_2026}. Other methods use human demonstrations directly for policy learning, through domain adaptation~\cite{punamiya2025egobridge}, pretraining or co-training~\cite{kareer2025egomimic,yuan_motiontrans_2025,yang_egovla_2025,zheng2026egoscale,shi2026egohumanoid,zhu_emma_2025,bi2026hrdt,wei2026psi0}, and transfer without additional target-robot demonstrations~\cite{xu2026hommi}.
Much of this literature emphasizes transferring human experience to target robot embodiments, while EgoX focuses on collecting and jointly learning from egocentric demonstrations across human and heterogeneous robot embodiments including simulated robots, and investigates how this shared training benefits manipulation across diverse platforms.

\parab{World Action Models and Embodiment Modeling.}
World models learn to predict how the world evolves under actions~\cite{assran2025vjepa2,goswami_world_2025,gao2026dreamdojo}.
World action models couple such prediction with policy learning by jointly predicting actions and future observations, with targets including pixels, pretrained visual features, and 3D motion flow~\cite{li_egowam_2026}, embeddings from frozen or separately pretrained encoders~\cite{zheng2025flare,luo2026beingh07}, and future tactile signals or tactile latents~\cite{niu2026learning,lou2026dreamtac}.
These works show that predicting future observations provides a useful training signal for policy learning.
Other methods model the embodiment itself, either by predicting its dynamics or kinematics~\cite{fu_metis_2025,patel2025getzero}, or by encoding its structure through body graphs, canonical morphology representations, or embodiment-specific functional components in action generation~\cite{sferrazza2024body,wei2026onehand,li2026xdiffvla}.
EgoX combines the ingredients of world action models and embodiment modeling through embodiment dreaming: alongside actions, MED predicts future embodiment latents from egocentric visual features, robot joint states, and end-effector joint states using exponential-moving-average copies of its own modality encoders. The objective helps the shared trunk learn embodiment-aware dynamics while leaving the deployment-time action policy unchanged.

\begin{figure}[t]
    \centering
    \includegraphics[width=\linewidth]{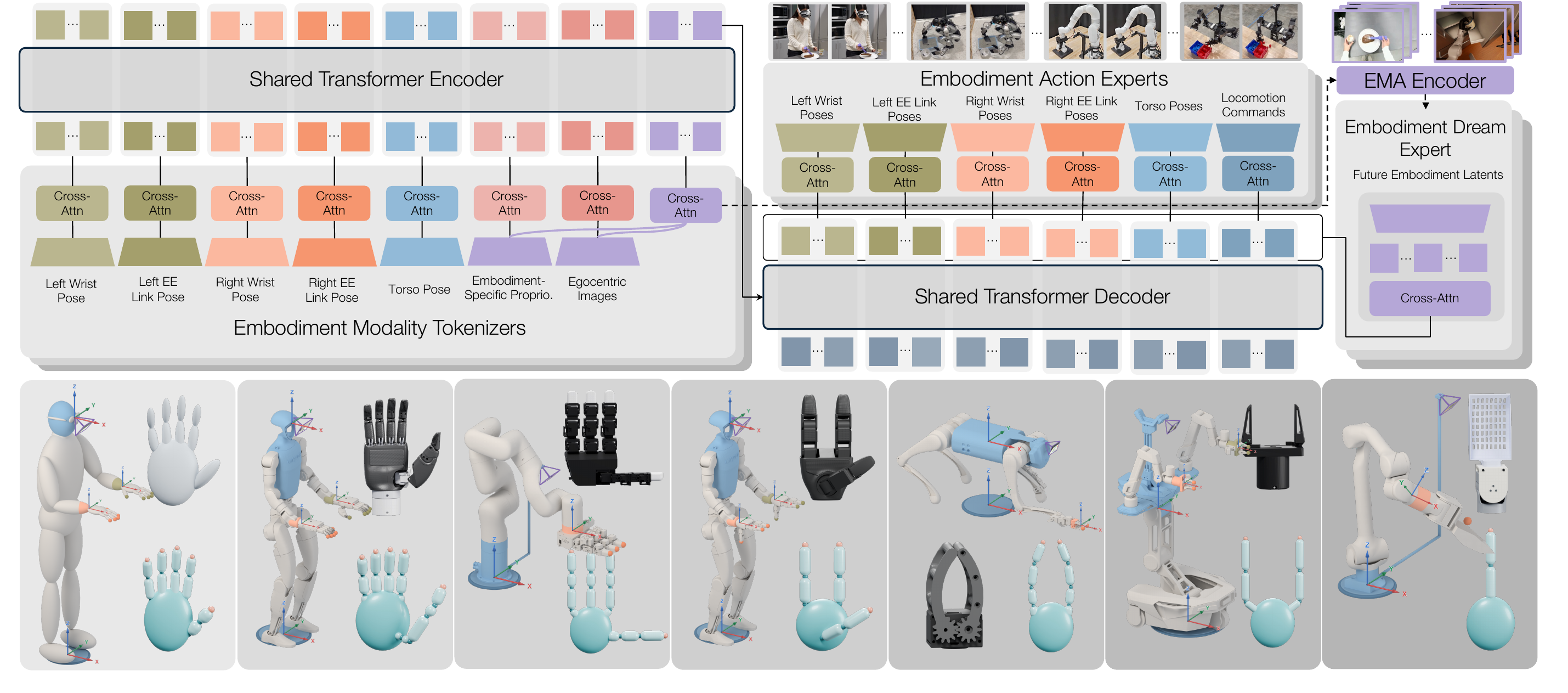}
    \caption{\small
    \textbf{MED architecture and EgoX embodiments.}
    \textit{Top:} embodiment-specific tokenizers encode each available observation modality
    (\modcolor{tokIM}{egocentric images}, \modcolor{tokPP}{embodiment-specific proprioception},
    \modcolor{tokLW}{left}/\modcolor{tokRW}{right} wrist poses,
    \modcolor{tokLE}{left}/\modcolor{tokRE}{right} end-effector keypoint positions, and \modcolor{tokTO}{torso poses})
    into tokens processed by a shared Transformer encoder--decoder trunk.
    A dedicated action expert decodes each action modality into an action chunk, while a
    dream expert predicts \modcolor{tokPU}{future embodiment latents}, supervised
    by EMA copies of the feature encoders and embodiment tokenizer.
    \textit{Bottom:} the embodiments used in EgoX, from left to right: \texttt{Human-Dex5}, \texttt{G1-Dex5},
    \texttt{xArm7-Dex4}, \texttt{G1-Sim-Dex3}, \texttt{LocoMan-Dex2}, \texttt{R1Lite-Gripper},
    and \texttt{Z1-Tool}. Clay models highlight body and wrist regions and end-effector keypoints
    in the corresponding modality colors; purple frustums mark the observation cameras.
    Frames are drawn schematically (defined in Sec.~\ref{sec:method_data}): torso poses use the (pseudo-)root frame,
    wrist poses and targets use the egocentric frame, and end-effector keypoint positions use
    the corresponding wrist frame. Insets show each end-effector alongside its keypoint representation.
    }
    \label{fig:architecture}
\end{figure}

\section{EgoX: Egocentric Cross-Embodiment Manipulation}
\label{sec:methodology}
\method unifies heterogeneous data into a shared modality structure, with coordinate conventions chosen for each modality (Sec.~\ref{sec:method_data}). To handle diverse sensing and control interfaces with different robots while facilitating positive transfer, \method leverages MED (Fig.~\ref{fig:architecture}), where embodiment-specific tokenizers encode each observation modality for a shared Transformer backbone, and modular action experts decode its outputs into each action modality (Sec.~\ref{sec:method_mxt}). MED also uses \emph{embodiment dreaming} to encourage the Transformer to learn each embodiment's dynamics (Sec.~\ref{sec:method_embodiment_dreaming}).
Sec.~\ref{sec:method_training} covers cross-embodiment pretraining and target-task fine-tuning.

\subsection{Egocentric Cross-Embodiment Data Representation}\label{sec:method_data}
Although embodiments differ substantially as whole bodies, many of their parts play corresponding roles. We exploit this by decomposing each embodiment into anatomically or functionally corresponding modalities and aligning their observations and actions in a shared structure.
Formally, \method collects
demonstrations from a set of embodiments $\mathcal{E}$ that includes both humans and
robots. For each embodiment $e \in \mathcal{E}$, the dataset
\begin{equation}
\mathcal{D}_{e} = \big\{ \tau^{e}_{i} \big\}_{i=1}^{N_e},
\qquad
\tau^{e}_{i} = \big( \bm{o}^{e}_{i,t},\, \bm{a}^{e}_{i,t} \big)_{t=1}^{T^{e}_{i}},
\end{equation}
contains $N_e$ trajectories. Trajectory $\tau^{e}_{i}$ has length $T^{e}_{i}$. The terms $\bm{o}^{e}_{i,t}$ and $\bm{a}^{e}_{i,t}$ are the observations and actions at time step $t$. Both are tuples of named modalities, defined below.
When the trajectory is clear from context, we omit the index $i$.

\begin{wrapfigure}[38]{R}{0.5\linewidth}
  \centering
  \includegraphics[width=\linewidth]{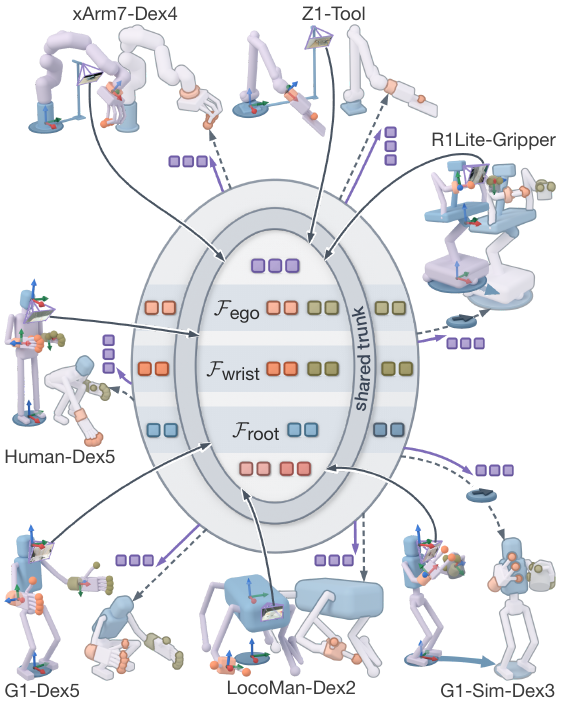}
  \caption{\small
  \textbf{Unified egocentric frames and shared token spaces.}
  Solid bodies show the current time step, with coordinate frames, camera views, and pale purple parts for any embodiment-specific proprioception. Translucent bodies show the pose one action chunk later.
  Solid, dashed, and purple arrows denote MED's tokenizers, action experts, and dream experts.
  The tokenizers map observations into one inner token space: \modcolor{tokRW}{right}/\modcolor{tokLW}{left} wrists at $\mathcal{F}_{\text{ego}}$, \modcolor{tokRE}{right}/\modcolor{tokLE}{left} end-effector links at their $\mathcal{F}_{\text{wrist}}$, the \modcolor{tokTO}{torso} at $\mathcal{F}_{\text{root}}$, and frame-free \modcolor{tokPP}{proprioception}, \modcolor{tokIM}{image}, and \modcolor{tokPU}{embodiment} tokens.
  The shared trunk fuses all tokens into outer action tokens. The action experts decode the action chunk, including any \modcolor{tokLO}{velocity commands} at $\mathcal{F}_{\text{root}}$, drawn as arrowed discs, and the dream experts predict each body's training-only \modcolor{tokPU}{\mbox{future} latents}.
  }
  \label{fig:frames}
\end{wrapfigure}

\parab{Unified egocentric frame system.}
To retain a consistent spatial interpretation across embodiments, \method uses common, modality-specific reference-frame conventions (Figs.~\ref{fig:architecture} and~\ref{fig:frames}; Appendix~\ref{app:frames}). The root frame $\mathcal{F}_{\text{root}}$ is the ground projection of a reference body part, such as a humanoid's feet, a quadruped's trunk, or, for human data, the head at the start of the episode. The egocentric frame $\mathcal{F}_{\text{ego}}$ is attached to the egocentric main camera frame or the rigid body carrying the main camera, and a wrist frame $\mathcal{F}_{\text{wrist}}$ is attached to each wrist. \modcolor{tokTO}{Torso} poses and \modcolor{tokLO}{velocity commands} are expressed in $\mathcal{F}_{\text{root}}$, wrist poses and targets in $\mathcal{F}_{\text{ego}}$, and end-effector keypoint positions in the corresponding $\mathcal{F}_{\text{wrist}}$. At reset, $\mathcal{F}_{\text{root}}$ has the $x$-axis pointing forward toward the workspace, the $y$-axis left, and the $z$-axis upward.
Poses are converted from teleoperation, sensing, or control conventions into these frames before data saving and model inference, whereas joint states remain in each embodiment's joint space. %

\parab{Observations and action targets.} Using these frame conventions, \method maps each embodiment's available observations and action targets into a shared modality structure:
\begin{align}
\bm{o}^{e}_{t}
&:=
\{
\modcolor{tokIM}{\bm{I}^{e}_{t}},\,
\modcolor{tokPP}{\bm{s}^{e}_{t}},\,
\modcolor{tokTO}{\bm{x}^{e}_{\mathrm{torso},t}},\,
\modcolor{tokLW}{\bm{x}^{e}_{\mathrm{L},t}},\,
\modcolor{tokRW}{\bm{x}^{e}_{\mathrm{R},t}},\,
\modcolor{tokLE}{\bm{p}^{e}_{\mathrm{L},t}},\,
\modcolor{tokRE}{\bm{p}^{e}_{\mathrm{R},t}},
\ldots
\}, \\
\bm{a}^{e}_{t}
&:=
\{
\modcolor{tokTO}{\tilde{\bm{x}}^{e}_{\mathrm{torso},t}},\,
\modcolor{tokLW}{\tilde{\bm{x}}^{e}_{\mathrm{L},t}},\,
\modcolor{tokRW}{\tilde{\bm{x}}^{e}_{\mathrm{R},t}},\,
\modcolor{tokLE}{\tilde{\bm{p}}^{e}_{\mathrm{L},t}},\,
\modcolor{tokRE}{\tilde{\bm{p}}^{e}_{\mathrm{R},t}},
\ldots
\}.
\end{align}
Here, $\modcolor{tokIM}{\bm{I}}$ denotes the egocentric main-camera image, $\modcolor{tokTO}{\bm{x}_{\mathrm{torso}}}$ the torso pose, $\modcolor{tokLW}{\bm{x}_{\mathrm{L}}}$ and $\modcolor{tokRW}{\bm{x}_{\mathrm{R}}}$ the left and right wrist poses, and $\modcolor{tokLE}{\bm{p}_{\mathrm{L}}}$ and $\modcolor{tokRE}{\bm{p}_{\mathrm{R}}}$ the corresponding end-effector keypoint positions. The term $\modcolor{tokPP}{\bm{s}}$ collectively denotes the available embodiment-specific proprioceptive streams, mainly robot and end-effector joint states; each stream is tokenized separately. A tilde indicates a commanded target. Colors match the modalities in Figs.~\ref{fig:architecture} and~\ref{fig:frames}. Additional observations may include wrist-camera images and force feedback; additional actions may include \modcolor{tokLO}{velocity commands}.

End-effector keypoints are task-relevant 3D positions expressed in the corresponding wrist frame, such as the fingertips of human and robot hands and the contact points of grippers and tools (bottom of Fig.~\ref{fig:architecture}). By describing end-effector geometry rather than actuator joint coordinates, they provide comparable geometric descriptions for hands, grippers, and tools with different kinematics.

\parab{Shared modality vocabulary.}
Let $\mathcal{M} = \mathcal{M}_{\mathrm{obs}} \cup \mathcal{M}_{\mathrm{act}}$ be the modality vocabulary shared by all embodiments, indexed by $m$: each constituent stream of $\bm{o}^{e}_{t}$ and $\bm{a}^{e}_{t}$ above, with L and R counted separately, is one modality, written $\bm{o}^{e,m}_{t}$ or $\bm{a}^{e,m}_{t}$. Not every embodiment provides every modality, so $\mathcal{M}^{e}_{\mathrm{obs}} \subseteq \mathcal{M}_{\mathrm{obs}}$ and $\mathcal{M}^{e}_{\mathrm{act}} \subseteq \mathcal{M}_{\mathrm{act}}$ denote the modalities available for embodiment $e$. Missing observation modalities are omitted or replaced by zero tokens, and missing action modalities are excluded from the loss (Appendix~\ref{sec:app_missing_modalities}).
Figure~\ref{fig:frames} illustrates how corresponding torso, wrist, and end-effector keypoint modalities across embodiments are expressed in their respective reference frames. Embodiment-specific tokenizers map these modalities, together with the available visual and proprioceptive observations, into a shared token space.

\subsection{Modular Embodiment Dreamer} \label{sec:method_mxt}
To learn from heterogeneous human and robot demonstrations, EgoX uses MED (Modular Embodiment Dreamer).
Building on prior modular tokenization and action decoding~\cite{niu2025human2locoman},
we introduce finer-grained interfaces for left/right wrist poses and end-effector keypoints.
Within each embodiment, each available observation modality
has a dedicated tokenizer and each action modality has a dedicated expert, connected through a
shared encoder--decoder Transformer trunk.
During training, an additional embodiment-specific \modcolor{tokPU}{dream expert} $\modcolor{tokPU}{\psi^{e}_{\mathrm{dream}}}$
predicts \modcolor{tokPU}{future embodiment latents} from the shared trunk (Sec.~\ref{sec:method_embodiment_dreaming}).

\parab{Architecture overview.}
For each embodiment $e \in \mathcal{E}$, MED's action policy is the composition
\begin{equation}
\ifdim\displaywidth<.75\textwidth
\prevgraf=\numexpr\prevgraf+2\relax
\begin{aligned}
\pi^{e} &= \Psi^{e} \circ G_{\theta} \circ \Phi^{e},\\
\Phi^{e} &= \{\phi^{e}_{m}\}_{m \in \mathcal{M}^{e}_{\mathrm{obs}}} \cup \{\modcolor{tokPU}{\phi^{e}_{\mathrm{dream}}}\},\\
\Psi^{e} &= \{\psi^{e}_{m}\}_{m \in \mathcal{M}^{e}_{\mathrm{act}}},
\end{aligned}
\else
\pi^{e} = \Psi^{e} \circ G_{\theta} \circ \Phi^{e},
\qquad
\Phi^{e} = \{\phi^{e}_{m}\}_{m \in \mathcal{M}^{e}_{\mathrm{obs}}} \cup \{\modcolor{tokPU}{\phi^{e}_{\mathrm{dream}}}\},
\qquad
\Psi^{e} = \{\psi^{e}_{m}\}_{m \in \mathcal{M}^{e}_{\mathrm{act}}},
\fi
\end{equation}
where $\phi^{e}_{m}$ is the tokenizer of observation modality $m$, $\psi^{e}_{m}$ is the action
expert of action modality $m$, and $G_{\theta}$ is the shared trunk. Here, $\Phi^{e}$ and $\Psi^{e}$
denote the collective encoding and decoding operations of the listed modules. The \modcolor{tokPU}{embodiment
tokenizer} $\modcolor{tokPU}{\phi^{e}_{\mathrm{dream}}}$ (Sec.~\ref{sec:method_embodiment_dreaming}) aggregates features of the
current egocentric image and proprioception into \modcolor{tokPU}{embodiment tokens}, which are concatenated with the
modality tokens before entering the shared trunk. Modality tokenizers and action experts are
instantiated only for the modalities that $e$ provides. The shared modality vocabulary preserves
anatomical or functional correspondences across embodiments, while embodiment-specific
tokenizers and experts accommodate differences in sensing and control. The shared trunk serves
as the main pathway for cross-embodiment learning. Given an observation
$\bm{o}^{e}_{t} = (\bm{o}^{e,m}_{t})_{m \in \mathcal{M}^{e}_{\mathrm{obs}}}$, the policy predicts
an action chunk, i.e., the next $H$ action targets of every available action modality:
\begin{equation}
\hat{\bm{A}}^{e}_{t} = \pi^{e}\big(\bm{o}^{e}_{t}\big)
= \big(\hat{\bm{A}}^{e,m}_{t}\big)_{m \in \mathcal{M}^{e}_{\mathrm{act}}},
\qquad
\hat{\bm{A}}^{e,m}_{t} = \big(\hat{\bm{a}}^{e,m}_{t+k}\big)_{k=0}^{H-1} \in \mathbb{R}^{H \times d_m^{e}},
\end{equation}
where $d_m^{e}$ is the dimension of action modality $m$ for embodiment $e$. Appendix~\ref{sec:app_architecture} provides details on each module.

\subsection{Embodiment Dreaming} \label{sec:method_embodiment_dreaming}
Behavioral cloning supervises the policy through demonstrated actions, without explicitly requiring
it to anticipate how the embodiment's state evolves. MED adds \emph{\modcolor{tokPU}{embodiment dreaming}},
which predicts compact \modcolor{tokPU}{future embodiment latents} to encourage the shared trunk to
capture embodiment-specific dynamics without reconstructing high-dimensional observations.
Following the joint-embedding predictive architecture (JEPA)~\cite{assran2025vjepa2}, the student \modcolor{tokPU}{embodiment tokenizer} serves as the context encoder and its EMA teacher
as the target encoder, encoding future egocentric images and embodiment-specific proprioception into these latents.
Thus, no separately pretrained embodiment encoder is needed, and \modcolor{tokPU}{dreaming} is trained jointly with action prediction.

\parab{Embodiment latent.}
For embodiment $e$, let $\mathcal{M}^{e}_{\mathrm{dream}} \subseteq \mathcal{M}^{e}_{\mathrm{obs}}$
be the observation modalities used as dreaming sources: the egocentric image, robot joint
states, and end-effector joint states, whichever of them $e$ provides. Robot joint
states are not part of the action space, but they carry embodiment-specific state that is
informative about body dynamics. Let $f^{e}_{m}$ denote the feature-encoding component of modality
tokenizer $\phi^{e}_{m}$, including positional embeddings, before modality-specific aggregation.
The \modcolor{tokPU}{embodiment tokenizer} $\modcolor{tokPU}{\phi^{e}_{\mathrm{dream}}}$ aggregates these features by
cross-attention from $L_{\mathrm{dream}}$ learnable queries into
$L_{\mathrm{dream}}$ latent tokens of dimension $d$ ($L_{\mathrm{dream}} = 8$ in our experiments).
These tokens are retained without pooling. EMA copies of
$f^{e}_{m}$ and $\modcolor{tokPU}{\phi^{e}_{\mathrm{dream}}}$, marked by a bar, summarize future observation features
to provide targets for the \modcolor{tokPU}{dream expert} at offset $k$:
\begin{equation}
\modcolor{tokPU}{\bm{Y}^{e}_{t+k}}
=
\mathrm{sg}\Big[\,
\modcolor{tokPU}{\bar{\phi}^{e}_{\mathrm{dream}}}\Big(
\big[\, \bar{f}^{e}_{m}\big(\bm{o}^{e,m}_{t+k}\big) \,\big]_{m \in \mathcal{M}^{e}_{\mathrm{dream}}}
\Big)
\Big].
\label{eq:emb_dream_target}
\end{equation}
Here $\modcolor{tokPU}{\bm{Y}^{e}_{t+k}} = [\modcolor{tokPU}{\bm{y}^{e}_{t+k,j}}]_{j=1}^{L_{\mathrm{dream}}}
\in \mathbb{R}^{L_{\mathrm{dream}} \times d}$ is the sequence of target tokens, and
$\mathrm{sg}[\cdot]$ is the stop-gradient operator.

\parab{EMA target update.}
Each teacher in Eq.~\eqref{eq:emb_dream_target} tracks its corresponding student:
$\modcolor{tokPU}{\phi^{e}_{\mathrm{dream}}}$ or $f^{e}_{m}$ for $m \in \mathcal{M}^{e}_{\mathrm{dream}}$.
For student parameters $w$ and their teacher counterparts $\bar{w}$, after each optimizer step,
\begin{equation}
\bar{w} \leftarrow \alpha\, \bar{w} + (1 - \alpha)\, w,
\qquad
\alpha \in (0, 1),
\label{eq:emb_ema}
\end{equation}
where $\alpha$ is the EMA decay. As in JEPA, the stop-gradient prevents the \modcolor{tokPU}{dreaming loss} from updating
the target, and the EMA update makes the target evolve slowly.

\parab{Dream expert.}
The \modcolor{tokPU}{dream expert} predicts these target sequences at a set of future offsets $\mathcal{K}$:
\begin{equation}
\big(\modcolor{tokPU}{\hat{\bm{Y}}^{e}_{t+k}}\big)_{k \in \mathcal{K}}
=
\big(\modcolor{tokPU}{\psi^{e}_{\mathrm{dream}}} \circ G_{\theta} \circ \Phi^{e}\big)\big(\bm{o}^{e}_{t}\big).
\end{equation}
Dreaming targets are sampled from future observations by the same data pipeline as action chunks.

\parab{Latent prediction loss.}
Each predicted token is supervised against its corresponding target token with a cosine
alignment loss and an optional magnitude loss, averaged over future offsets and token indices:
\begin{align}
\modcolor{tokPU}{\mathcal{L}^{e}_{\mathrm{dream}}}
&=
\frac{1}{|\mathcal{K}| L_{\mathrm{dream}}}
\sum_{k \in \mathcal{K}}
\sum_{j=1}^{L_{\mathrm{dream}}}
\Big[
1 - \cos\big(\modcolor{tokPU}{\hat{\bm{y}}^{e}_{t+k,j}},\, \modcolor{tokPU}{\bm{y}^{e}_{t+k,j}}\big)
\notag \\
&\qquad
+ \beta\, \ell_{\text{smooth-L1}}\big(
\|\modcolor{tokPU}{\hat{\bm{y}}^{e}_{t+k,j}}\|_2,\,
\|\modcolor{tokPU}{\bm{y}^{e}_{t+k,j}}\|_2
\big)
\Big],
\label{eq:emb_dream_loss}
\end{align}
where $\beta \ge 0$ weights the magnitude term.

\subsection{Cross-Embodiment Pretraining and Target Fine-Tuning} \label{sec:method_training}
EgoX first pretrains MED on a mixture of human and robot embodiments with action prediction and
\modcolor{tokPU}{embodiment dreaming}, and then fine-tunes it on demonstrations of the target embodiment. %

\parab{Training objective.}
For a mini-batch from embodiment $e$, the action loss is a weighted sum of chunk losses over
the available action modalities,
\begin{equation}
\mathcal{L}^{e}_{\mathrm{act}}
=
\sum_{m \in \mathcal{M}^{e}_{\mathrm{act}}}
\lambda^{e}_{m}\,
\ell_{\mathrm{act}}\big(\hat{\bm{A}}^{e,m},\, \bm{A}^{e,m}\big),
\label{eq:act_loss}
\end{equation}
where $\hat{\bm{A}}^{e,m}$ and $\bm{A}^{e,m}$ collect the predicted and demonstrated action chunks
in the mini-batch, and $\lambda^{e}_{m}$ weights action modality $m$. The loss $\ell_{\mathrm{act}}$
averages elementwise smooth L1 errors over all valid chunk steps and action dimensions in the
mini-batch, using standardized actions (Appendix~\ref{sec:app_algorithm}).
In both Eq.~\eqref{eq:act_loss} and Eq.~\eqref{eq:emb_dream_loss}, chunk steps and dreaming
offsets beyond the end of the trajectory are padded and excluded from the average. The objective
for embodiment $e$ on a dataset $\mathcal{D}$ is
\begin{equation}
\mathcal{L}^{e}(\mathcal{D})
=
\mathbb{E}_{\mathcal{D}}
\big[\,
\mathcal{L}^{e}_{\mathrm{act}}
+
\lambda^{e}_{\mathrm{dream}}\, \modcolor{tokPU}{\mathcal{L}^{e}_{\mathrm{dream}}}
\,\big],
\label{eq:emb_loss}
\end{equation}
where $\mathbb{E}_{\mathcal{D}}$ denotes expectation over mini-batches sampled from
$\mathcal{D}$, and $\lambda^{e}_{\mathrm{dream}} \ge 0$ weights \modcolor{tokPU}{embodiment dreaming}.

\parab{Pretraining.}
Let $\mathcal{E}_{\mathrm{pre}} \subseteq \mathcal{E}$ be the pretraining embodiments, including
humans and multiple robots. Each optimizer step accumulates gradients from a mini-batch from
every pretraining embodiment, minimizing
\begin{equation}
\mathcal{L}_{\mathrm{pre}}
=
\sum_{e \in \mathcal{E}_{\mathrm{pre}}}
\mathcal{L}^{e}(\mathcal{D}_{e}).
\label{eq:pretrain_loss}
\end{equation}
Demonstrations of the target task on the target embodiment are excluded from pretraining.
The shared trunk $G_{\theta}$ receives gradients from all embodiments, whereas the
embodiment-specific modules of $e$ are updated only by $\mathcal{D}_{e}$.

\parab{Fine-tuning.}
Given a target embodiment $e^{\star}$ and a target-task dataset
$\mathcal{D}^{\mathrm{task}}_{e^{\star}}$, we initialize MED from the pretrained checkpoint
and continue training with both action prediction and \modcolor{tokPU}{embodiment dreaming}:
\begin{equation}
\mathcal{L}_{\mathrm{ft}}
=
\mathcal{L}^{e^{\star}}\big(\mathcal{D}^{\mathrm{task}}_{e^{\star}}\big).
\label{eq:finetune_loss}
\end{equation}
We transfer the shared components and any pretrained target-embodiment modules, initializing
new modules where needed (Appendix~\ref{sec:app_algorithm}). Fine-tuning adapts the shared
representation and target-specific modules to the target task, building on the manipulation
priors learned through cross-embodiment pretraining.

\parab{Inference.}
At deployment, MED retains the modality tokenizers and student \modcolor{tokPU}{embodiment tokenizer}
$\modcolor{tokPU}{\phi^{e^{\star}}_{\mathrm{dream}}}$. The latter aggregates current observation features into
\modcolor{tokPU}{embodiment tokens}, which are concatenated with the modality tokens and passed to the shared
trunk. The action experts decode action chunks $\hat{\bm{A}}^{e^{\star}}_{t}$ for low-level
control. The \modcolor{tokPU}{dream expert} $\modcolor{tokPU}{\psi^{e^{\star}}_{\mathrm{dream}}}$ and EMA teachers are used only
during training; inference requires neither future-latent prediction nor EMA target computation.
Algorithm~\ref{alg:egox_training} in Appendix~\ref{sec:app_algorithm} summarizes the training procedure.

\section{Experiments and Results}
\label{sec:experiments}
\subsection{Experimental Setup}

\parab{Tasks.}
We evaluate policies on 13 task--embodiment pairs spanning five manipulation tasks and six robot embodiments, covering long-horizon loco-manipulation, semantic object selection, precise alignment, tool use, and contact-rich interaction.
In \texttt{Kitchen-Bread} (Sauce Spreading), the robot retrieves sauce containers from a fridge or tabletop, picks up a spreading tool, dips it into the sauce, and spreads the sauce onto bread.
In \texttt{Kitchen-Cleanup} (Tool Organization), the robot picks up tools from a tabletop and places them into a container.
In \texttt{Sort} (Block Sorting), the robot sorts blocks into containers of the corresponding colors.
In \texttt{Insert} (T-shaped Block Insertion), the robot picks up a T-shaped block, aligns it with its base, and inserts it.
In \texttt{Scoop} (Cat Litter Scooping), the robot scoops cat litter from a litter box into a disposal bin, then returns the scoop to the edge of the box.
Figs.~\ref{fig:task_scoop}--\ref{fig:task_kitchen_cleanup} in Appendix~\ref{app:evaluation_protocol} show example rollouts of each task across the evaluated embodiments.

\parab{Embodiments.}
Our evaluation covers six robot embodiments (five physical robots and one simulated humanoid), spanning diverse morphologies, mobility configurations, end-effectors, and unimanual or bimanual operation. \texttt{G1-Dex5} is a bimanual Unitree G1 humanoid equipped with five-finger Inspire RH56F1 dexterous hands. \texttt{xArm7-Dex4} is a tabletop xArm7 equipped with a four-finger LEAP Hand~\cite{shaw2023leap}; \texttt{G1-Sim-Dex3} is a simulated Unitree G1 equipped with three-finger Dex3-1 dexterous hands. \texttt{LocoMan-Dex2}~\cite{lin2024locoman} is a quadruped with two-finger grippers mounted on its front legs, evaluated in unimanual mode. \texttt{R1Lite-Gripper} is a Galaxea R1 Lite wheeled bimanual mobile manipulator with parallel-jaw grippers. \texttt{Z1-Tool} is a tabletop Unitree Z1 arm with a fixed, non-actuated scoop. Egocentric human demonstrations (\texttt{Human-Dex5}) provide an additional embodiment for cross-embodiment pretraining. Together, these platforms span dexterous hands, grippers, and fixed tools across diverse robot morphologies and mobility configurations.

\parab{Baselines and training settings.}
We compare MED with ACT~\cite{zhao2023learning} and $\pi_{0.5}$~\cite{intelligence2025pi_}, and evaluate MED both with and without cross-embodiment pretraining. ACT is trained on demonstrations from the target embodiment and task, while $\pi_{0.5}$ is fine-tuned from its pretrained checkpoint on the same demonstrations (Appendix~\ref{app:baselines}). \emph{MED without pretraining} is trained from scratch on these target-task demonstrations. \emph{MED with pretraining} is first trained on a mixture of human, physical-robot, and simulated demonstrations, excluding demonstrations of the target task on the target embodiment, and then fine-tuned on those held-out demonstrations following Sec.~\ref{sec:method_training}. Both MED variants use the same downstream policy architecture and training objective, including \modcolor{tokPU}{embodiment dreaming}, to assess the benefit of cross-embodiment pretraining. We additionally ablate the pretraining composition, \modcolor{tokPU}{embodiment dreaming}, and modality-level modularity.

\parab{Evaluation settings and metrics.}
For each task--embodiment pair, we assess policies under in-distribution (ID) and out-of-distribution (OOD) conditions, defined relative to the target-task demonstrations of the evaluated embodiment. ID settings use similar objects, initial configurations, and environmental conditions to those in these demonstrations, while OOD settings introduce held-out variations in object instances, appearances, poses, or scene configurations. Similar conditions may appear in other embodiments' pretraining demonstrations, allowing us to assess cross-embodiment transfer. For each method, we run 20 trials (13 ID and 7 OOD) per real-world task--embodiment pair and 50 trials (32 ID and 18 OOD) per simulated pair, using the same trial configurations across methods. We report the \textit{mean normalized task progress score} separately for ID and OOD conditions. Each trial's score credits intermediate stages and adds a one-point bonus for successful completion, normalized by the task's maximum score including the bonus. Task-specific evaluation and scoring criteria are detailed in Appendix~\ref{app:evaluation_protocol}; success rates on the same trials are reported in Appendix~\ref{app:success_rate}.

\subsection{Experimental Evaluation}

Our evaluation aims to address the following questions:
(1) Does \textit{egocentric cross-embodiment pretraining} improve downstream manipulation compared to training from scratch on target-embodiment data? %
(2) How does \textit{pretraining composition} affect transfer? %
(3) Does \textit{embodiment dreaming} improve cross-embodiment learning?
(4) Does \textit{modular design} enhance cross-embodiment transfer?
(5) How do cross-embodiment pretraining and embodiment dreaming affect latent geometry across embodiments?

\begin{figure*}[!t]
    \centering
    \includegraphics[width=\linewidth]{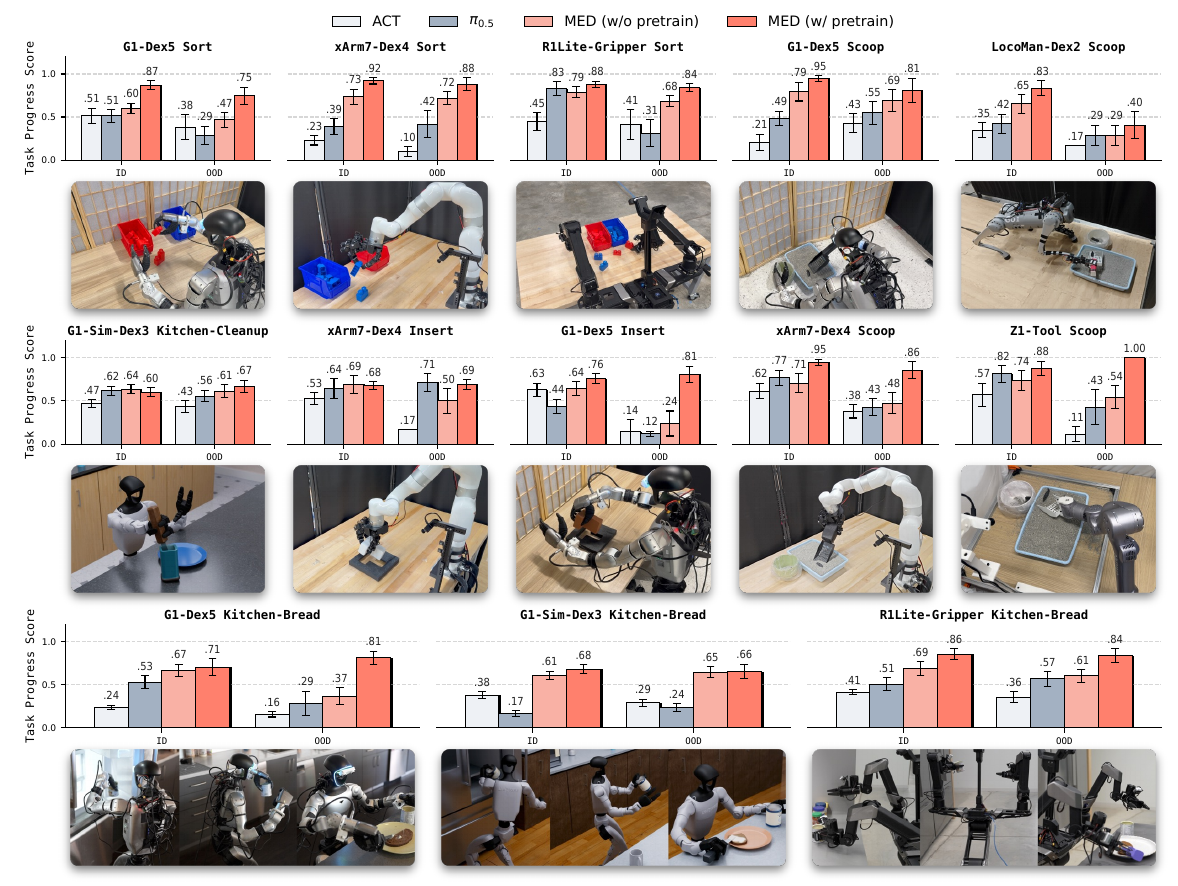}%
    \vspace{-0.2cm}
    \caption{\small
    \textbf{Manipulation performance across tasks and embodiments.}
    Normalized task progress of ACT, $\pi_{0.5}$, MED without pretraining, and MED with cross-embodiment pretraining across 13 task--embodiment pairs spanning five tasks and six embodiments (five physical robots and the simulated \texttt{G1-Sim-Dex3}). Each panel reports ID and OOD performance. Success rates: Fig.~\ref{fig:success_rate} in Appendix~\ref{app:success_rate}.}
    \label{fig:main_results}
\end{figure*}

\parab{Does egocentric cross-embodiment pretraining outperform training from scratch?} Cross-embodiment pretraining raises MED's mean normalized task progress from 0.63 to 0.80 across the 13 task--embodiment pairs (Fig.~\ref{fig:main_results}), pooling ID and OOD trials within each pair before averaging across pairs. Improvements occur on 11 of 13 pairs under ID evaluation and all 13 pairs under OOD evaluation. The gains are particularly pronounced under OOD conditions, where mean task progress increases from 0.53 to 0.77, with large improvements on \texttt{G1-Dex5 Insert} and \texttt{Z1-Tool Scoop}. Among all evaluated methods, pretrained MED achieves the highest average task progress across the 13 task--embodiment pairs and ranks first on 12 of them. These results suggest that learning from heterogeneous human and robot experience improves downstream manipulation, particularly under conditions absent from the target-task demonstrations of the evaluated embodiment.

\begin{figure*}[!ht]
    \centering
    \begin{minipage}[b]{0.49\linewidth}
        \centering
        \includegraphics[width=\linewidth]{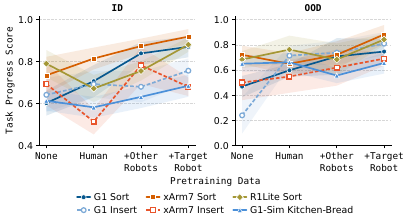}
    \end{minipage}\hfill
    \begin{minipage}[b]{0.49\linewidth}
        \centering
        \includegraphics[width=\linewidth]{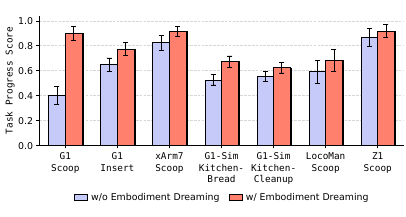}
    \end{minipage}

    \vspace{-0.3cm}
    \begin{minipage}[t]{0.49\linewidth}
        \caption{\raggedright\small\textbf{Pretraining data composition ablation.} Task progress as pretraining data is added: human, other robots, then the target robot's other tasks. ID and OOD use different y ranges. Labels omit end effectors. Shading: $\pm$1 SEM. Success rates: Table~\ref{tab:sr_composition} in Appendix~\ref{app:success_rate}.}
        \label{fig:ablation_composition}
    \end{minipage}\hfill
    \begin{minipage}[t]{0.49\linewidth}
        \caption{\raggedright\small\textbf{Embodiment dreaming ablation.} Task progress with and without embodiment dreaming across seven task--embodiment pairs. ID and OOD trials are pooled within each pair. Labels omit end effectors. Error bars: $\pm$1 SEM. Success rates: Table~\ref{tab:sr_dreaming} in Appendix~\ref{app:success_rate}.}
        \label{fig:ablation_dreaming}
    \end{minipage}
    \vspace{-0.3cm}
\end{figure*}

\parab{How does pretraining composition affect transfer?}
We next study how the composition of the pretraining mixture affects downstream performance. As shown in Fig.~\ref{fig:ablation_composition}, we compare four settings: no pretraining, human-only pretraining, human plus other-robot pretraining, and full cross-embodiment pretraining. Human-only pretraining uses egocentric human demonstrations; human plus other-robot pretraining additionally incorporates demonstrations from robot embodiments other than the target robot; and full cross-embodiment pretraining further includes demonstrations from the target embodiment on non-target tasks. All variants use the same target-embodiment, target-task demonstrations for downstream training. We evaluate this ablation across both real-robot and simulation settings, covering \texttt{Sort} on \texttt{R1Lite-Gripper}, \texttt{Sort} and \texttt{Insert} on \texttt{G1-Dex5} and \texttt{xArm7-Dex4}, as well as \texttt{Kitchen-Bread} on simulated \texttt{G1-Sim-Dex3}.

Every pretraining mixture improves average OOD performance over no pretraining, and the average gains grow as the mixture broadens. Human-only pretraining already improves OOD performance on five of six pairs, most strongly on \texttt{G1-Dex5 Insert}, suggesting that egocentric human demonstrations can provide transferable manipulation priors despite the embodiment gap. Adding other robots' demonstrations raises ID performance on five of six pairs, and adding the target robot's demonstrations on other tasks improves OOD performance on every pair. Full cross-embodiment pretraining achieves the highest average task progress, improving over no pretraining by 0.12 on ID and 0.23 on OOD. These results show that combining human demonstrations, other robots' experience, and the target robot's other tasks improves downstream transfer.

\parab{Does \textit{embodiment dreaming} improve cross-embodiment learning?}
We compare MED with MXT-FG, our fine-grained (FG) adaptation of the Modularized Cross-Embodiment Transformer (MXT)~\cite{niu2025human2locoman}. MXT-FG retains MED's modality tokenizers and action experts but omits the embodiment-dreaming components and training objective during both pretraining and fine-tuning (Appendix~\ref{sec:app_algorithm}). MED improves average task progress from 0.63 to 0.78 across seven task--embodiment settings spanning five embodiments (Fig.~\ref{fig:ablation_dreaming}). These settings include \texttt{Scoop} on four embodiments (\texttt{G1-Dex5}, \texttt{xArm7-Dex4}, \texttt{LocoMan-Dex2}, and \texttt{Z1-Tool}), \texttt{Insert} on \texttt{G1-Dex5}, and both \texttt{Kitchen-Bread} and \texttt{Kitchen-Cleanup} on \texttt{G1-Sim-Dex3}. Task progress improves in all seven settings, with the largest gain on \texttt{G1-Dex5 Scoop}, supporting the benefit of the embodiment-dreaming components for downstream manipulation.

\parab{Does \textit{modular design} enhance cross-embodiment transfer?}\par\nobreak
\begin{wrapfigure}{r}{0.48\textwidth}
  \vspace{-\intextsep}
  \centering
  \includegraphics[width=\linewidth]{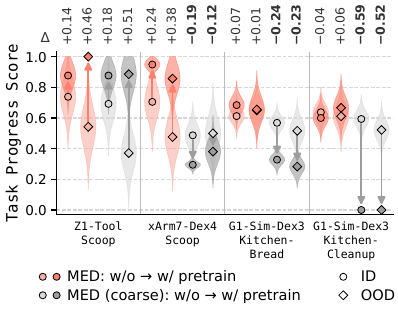}
  \caption{\raggedright\small\textbf{Coarse ablation.} Arrows go from w/o to w/ pretraining; violins show each mean's uncertainty (from its SEM). Success rates: Table~\ref{tab:sr_coarse} in Appendix~\ref{app:success_rate}.}
  \label{fig:ablation_coarse}
\end{wrapfigure}
We compare MED with a coarse variant that uses one tokenizer for all camera views, one for concatenated pose and proprioceptive streams, and one action expert for all action modalities within each embodiment. The image encoders, shared trunk, embodiment dreaming, training objective, and input/output token budgets remain unchanged. Fig.~\ref{fig:ablation_coarse} evaluates both variants with and without pretraining on \texttt{Z1-Tool Scoop}, \texttt{xArm7-Dex4 Scoop}, and \texttt{G1-Sim-Dex3 Kitchen-Bread} and \texttt{Kitchen-Cleanup}. These embodiments span progressively richer modality sets, from fixed-tool manipulation to dexterous-hand control and bimanual loco-manipulation with multiple camera views. After pretraining, modular MED matches or exceeds the coarse variant in every ID and OOD setting, with a small gap on \texttt{Z1-Tool} but larger gaps on \texttt{xArm7-Dex4} and \texttt{G1-Sim-Dex3}. Both variants benefit from pretraining on \texttt{Z1-Tool}. However, pretraining reduces the coarse variant's performance on \texttt{xArm7-Dex4 Scoop} and both simulated kitchen tasks, whereas modular MED improves in most settings and avoids these large drops.

On \texttt{Kitchen-Cleanup}, the pretrained coarse variant achieves zero task progress, while modular MED retains substantial performance. This task starts with locomotion, making its 3D velocity command critical despite its small share of the 51-dimensional action space. Dedicated action experts may help preserve this modality. Separate camera tokenizers may also help when wrist views are introduced during fine-tuning: MED retains its head-camera tokenizer and adds wrist-camera tokenizers, whereas the coarse variant jointly attends to all views. Added wrist views alone, however, do not explain the drops, which also occur on single-camera \texttt{xArm7-Dex4 Scoop}. Together, these results support modality-level modular design for effective cross-embodiment transfer, particularly for embodiments with diverse observation and action modalities.

\needspace{22\baselineskip}
\parab{How do cross-embodiment pretraining and embodiment dreaming affect latent geometry across embodiments?}\par\nobreak
\begin{wrapfigure}{r}{0.48\textwidth}
  \vspace{-\intextsep}
  \centering
  \includegraphics[width=\linewidth]{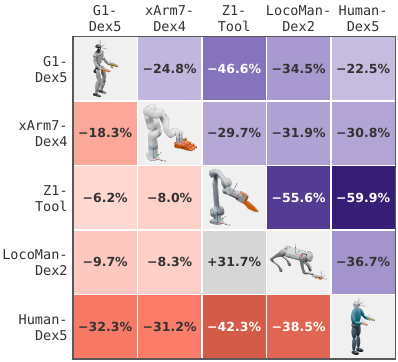}
  \caption{\raggedright\small GW change across embodiment pairs. \textbf{Pretraining}: coral, lower triangle; \textbf{\mbox{Embodiment dreaming}}: purple, upper triangle.}
  \label{fig:latent-gw}
\end{wrapfigure}
To assess pretraining, we compare MED with and without cross-embodiment pretraining. To assess the embodiment-dreaming components, we compare MED with MXT-FG, both pretrained and fine-tuned. For each comparison, we compute Gromov--Wasserstein (GW) distances between embodiment-specific distributions of shared-trunk latent representations extracted from \texttt{Scoop} demonstrations. Within each embodiment, we use matched demonstration samples across conditions, encoded by the corresponding checkpoints. For each sampled time step, we extract the right wrist-pose tokens from the shared-trunk output, before the action experts, and mean-pool them into one 512-dimensional latent vector. We use the right wrist pose because it is the most important action modality for \texttt{Scoop}, where the right end-effector is the only active end-effector shared across embodiments. GW compares the internal distance structures of these distributions using median-normalized within-distribution Euclidean distances in the original latent space. Lower GW distances indicate greater geometric similarity; we report percentage changes relative to the baseline in each comparison. To summarize each comparison, we first average GW over all 10 embodiment pairs for the baseline and for the compared model, and then report the percentage change from the baseline average to the compared model's average, rather than averaging the per-pair changes.

Fig.~\ref{fig:latent-gw} shows that cross-embodiment pretraining lowers GW for 9 of 10 embodiment pairs and reduces mean GW across the pairs by 21.30\%. In the embodiment-dreaming comparison, MED yields lower GW than MXT-FG for all 10 pairs, reducing mean GW across the pairs by 38.52\%. Both comparisons therefore show greater average geometric similarity among embodiment-specific latent representations of the same manipulation task.

\section{Conclusion}
\label{sec:conclusions}
We presented EgoX, a framework for collecting and learning from egocentric human, physical-robot, and simulated demonstrations through a shared modality structure. Treating these sources as embodiments on equal footing, EgoX pretrains on their combined experience and fine-tunes on each target embodiment and task. Its model, MED, connects each embodiment's modality-specific tokenizers and action experts through a shared Transformer trunk. Embodiment dreaming complements action supervision by predicting latent representations of future visual and proprioceptive observations, encouraging the trunk to capture how each embodiment's state evolves during interaction. Across 13 task--embodiment pairs spanning five tasks and six robot embodiments, cross-embodiment pretraining improves downstream manipulation, particularly under OOD conditions. Broader pretraining mixtures improve transfer, and embodiment dreaming further improves task performance and yields more similar cross-embodiment latent geometry on \texttt{Scoop}. Modality-level modularity is particularly beneficial for embodiments with diverse observation and action modalities, helping avoid the large performance drops observed when modalities are merged in the coarse variant. These findings support learning shared manipulation knowledge while preserving embodiment-specific sensing and control interfaces.

Several limitations merit further investigation. First, the demonstrations in our cross-embodiment experiments were collected by our team, which limits the data scale; performance with substantially larger training datasets remains untested. Rather than relying on widely used public robot manipulation datasets, many of which focus on arm-and-gripper systems, we prioritized demonstrations from a broader range of body morphologies and end-effectors to study transfer across substantially different embodiments. EgoX is nevertheless designed to accommodate existing datasets: its modality-structured representation can map their observations and actions into the shared structure, and its egocentric frame can be anchored to a known robot body rigidly connected to the camera, without requiring explicit camera calibration or camera-pose estimation. Public datasets could add complementary experience, but integrating them into EgoX and evaluating their benefit for cross-embodiment learning remain future work.

Second, our simulation experiments cover only one humanoid embodiment and two kitchen tasks, which limits the diversity of simulated experience. Third, adding a new embodiment to EgoX still requires manually specifying its modality correspondences and coordinate transforms, as well as integrating its control interface. Finally, although data availability varies substantially across embodiments, pretraining samples an equal-sized mini-batch from each. Samples from smaller datasets, such as \texttt{LocoMan-Dex2}, are therefore reused more often, which may increase overfitting and weaken shared pretraining. Important directions for future work include larger and more balanced datasets, broader simulation coverage, automated embodiment integration, and sampling strategies that account for each embodiment's data availability and diversity.

\bibliography{references}
\bibliographystyle{iclr2027_conference}

\clearpage
\appendix
\section{Appendix}
\label{sec:appendix}

\subsection{Handling Missing Modalities} \label{sec:app_missing_modalities}
Not every embodiment provides every modality in the shared vocabulary $\mathcal{M}$
(Sec.~\ref{sec:method_data}); for example, single-arm embodiments have no left wrist or left
end-effector link. \method handles such gaps so that token sequences remain semantically aligned
across embodiments. For each embodiment $e$, the modality indicator
\begin{equation}
\delta^{e}_{m} = \mathbb{I}\left[\, m \text{ is available for } e \,\right] \in \{0,1\}
\end{equation}
defines the available observation modalities
$\mathcal{M}^{e}_{\mathrm{obs}} = \{ m \in \mathcal{M}_{\mathrm{obs}} : \delta^{e}_{m} = 1 \}$,
and $\mathcal{M}^{e}_{\mathrm{act}}$ is defined analogously. Unavailable observation modalities are
omitted from the token sequence in our implementation (Appendix~\ref{sec:app_architecture}). Alternatively, missing modalities can be
kept in the sequence as zero tokens, so that embodiments share the same token layout. In this case,
the trunk's attention adds $\delta^{e}_{m}$ to the logits of keys from modality $m$, which
down-weights the zero tokens without excluding them. Unavailable action modalities are excluded
from the action loss (Sec.~\ref{sec:method_training}).

\subsection{Details on MED} \label{sec:app_architecture}
\parab{Modality tokenizers.}
Each tokenizer $\phi^{e}_{m}$ maps modality $m$ to a fixed number $L_m$ of $d$-dimensional tokens.
Observation modalities are defined at an anatomical or functional level, e.g., \modcolor{tokIM}{egocentric images},
\modcolor{tokTO}{torso} pose, \modcolor{tokLW}{left}/\modcolor{tokRW}{right} wrist poses, and \modcolor{tokLE}{left}/\modcolor{tokRE}{right} end-effector link poses, so embodiments with
different morphologies expose compatible interfaces whenever they share a modality. The
exception is \modcolor{tokPP}{embodiment-specific proprioception} (hereafter, proprioception), mainly robot and end-effector joint states,
whose dimensionality and meaning differ across embodiments. Image
modalities are encoded by a visual backbone $f_{\mathrm{vis}}$ shared across all embodiments,
followed by an embodiment-specific cross-attention aggregator $g^{e}_{m}$, so that
$\phi^{e}_{m} = g^{e}_{m} \circ f_{\mathrm{vis}}$. Pose and proprioception modalities are encoded by lightweight
MLPs followed by cross-attention aggregation. Only the available modalities are tokenized
(Appendix~\ref{sec:app_missing_modalities}). Their tokens are concatenated in a fixed modality
order, together with the current \modcolor{tokPU}{embodiment tokens}
$\modcolor{tokPU}{\bm{Z}^{e,\mathrm{dream}}_{t}}$ produced by the student \modcolor{tokPU}{embodiment tokenizer}
$\modcolor{tokPU}{\phi^{e}_{\mathrm{dream}}}$ from selected modality features (Sec.~\ref{sec:method_embodiment_dreaming}):
\begin{equation}
\begin{aligned}
\bm{Z}^{e}_{t} &= \big[\, (\bm{Z}^{e,m}_{t})_{m \in \mathcal{M}^{e}_{\mathrm{obs}}},\,
\modcolor{tokPU}{\bm{Z}^{e,\mathrm{dream}}_{t}} \,\big] \in \mathbb{R}^{L^{e} \times d},\\
L^{e} &= \textstyle\sum_{m \in \mathcal{M}^{e}_{\mathrm{obs}}} L_m + L_{\mathrm{dream}}.
\end{aligned}
\end{equation}
Here $\modcolor{tokPU}{\bm{Z}^{e,\mathrm{dream}}_{t}} \in \mathbb{R}^{L_{\mathrm{dream}} \times d}$ contains current
student tokens; the future EMA targets are defined separately in Eq.~\eqref{eq:emb_dream_target}.
Tokenizing per modality preserves each modality's semantic identity and avoids mixing
heterogeneous signals too early. The trunk therefore always consumes the same token interface,
while the tokenizers absorb embodiment-specific differences in dimensionality, camera
availability, and kinematic parameterization.

\parab{Shared Transformer trunk.}
The trunk $G_{\theta}$ is an encoder--decoder Transformer. The encoder applies self-attention over
$\bm{Z}^{e}_{t}$, and the decoder cross-attends from $N_q$ learnable query tokens
$\bm{Q} \in \mathbb{R}^{N_q \times d}$ to the encoder outputs:
\begin{equation}
\bm{U}^{e}_{t} = G_{\theta}\big(\bm{Z}^{e}_{t};\, \bm{Q}\big) \in \mathbb{R}^{N_q \times d}.
\end{equation}
Because every embodiment is presented through the same token interface, $G_{\theta}$ is trained
on all embodiments without requiring their raw observations or actions to share a
dimensionality. The trunk is therefore the main conduit for cross-embodiment transfer, while the
tokenizers and experts retain the embodiment-specific details needed for execution.

\parab{Modular action experts.}
For each available action modality $m \in \mathcal{M}^{e}_{\mathrm{act}}$, the action expert
decodes the corresponding action chunk from the trunk output:
\begin{equation}
\hat{\bm{A}}^{e,m}_{t} = \psi^{e}_{m}\big(\bm{U}^{e}_{t}\big) \in \mathbb{R}^{H \times d_m^{e}}.
\end{equation}
Action modalities are drawn from the shared vocabulary $\mathcal{M}_{\mathrm{act}}$, so
anatomically or functionally corresponding outputs occupy the same modality across embodiments.
For example, the wrist pose of a single-arm robot corresponds to the right wrist
pose of a bimanual embodiment. The experts remain embodiment-specific to handle differences in
control semantics and in which action dimensions an embodiment provides
(Appendix~\ref{sec:app_algorithm}), while the structure shared across embodiments is learned by the
trunk.

\subsection{Training Algorithm}
\label{sec:app_algorithm}
Algorithm~\ref{alg:egox_training} details the full EgoX training procedure, including cross-embodiment pretraining, target fine-tuning, \modcolor{tokPU}{embodiment dreaming}, and the EMA target update.
The rest of this section describes each step, and Table~\ref{tab:training_hparams} lists the
hyperparameters.
\begin{algorithm}[h]
\caption{EgoX pretraining and fine-tuning with embodiment dreaming}
\label{alg:egox_training}
\begin{algorithmic}[1]
\REQUIRE Datasets $\{\mathcal{D}_{e}\}_{e \in \mathcal{E}_{\mathrm{pre}}}$, target dataset $\mathcal{D}^{\mathrm{task}}_{e^{\star}}$
\REQUIRE Action horizon $H$, dreaming offsets $\mathcal{K}$, EMA decay $\alpha$, per-embodiment batch size $B$
\ENSURE Target policy $\pi^{e^{\star}} = \Psi^{e^{\star}} \circ G_{\theta} \circ \Phi^{e^{\star}}$
\STATE Initialize tokenizers, trunk $G_{\theta}$, action experts, and dream experts
\STATE Initialize each EMA teacher as a copy of its corresponding student
\FOR{each pretraining step}
    \FOR{each embodiment $e \in \mathcal{E}_{\mathrm{pre}}$}
        \STATE Sample $B$ time steps $t$ from $\mathcal{D}_{e}$, with observations $\bm{o}^{e}_{t}$, action chunks, and future observations $\bm{o}^{e}_{t+k}$ for $k \in \mathcal{K}$
        \STATE Predict action chunks $\hat{\bm{A}}^{e}_{t}$ and embodiment latent-token sequences $\modcolor{tokPU}{\hat{\bm{Y}}^{e}_{t+k}}$
        \STATE Compute dreaming target-token sequences $\modcolor{tokPU}{\bm{Y}^{e}_{t+k}}$ with Eq.~\eqref{eq:emb_dream_target}
        \STATE Compute $\mathcal{L}^{e}$ with Eq.~\eqref{eq:emb_loss} and accumulate its gradient
    \ENDFOR
    \STATE Update students with the accumulated gradient of $\mathcal{L}_{\mathrm{pre}}$ in Eq.~\eqref{eq:pretrain_loss}
    \STATE Update teachers with Eq.~\eqref{eq:emb_ema}
\ENDFOR
\STATE Initialize MED from the pretrained checkpoint
\FOR{each fine-tuning step}
    \STATE Sample a batch from $\mathcal{D}^{\mathrm{task}}_{e^{\star}}$
    \STATE Compute $\mathcal{L}_{\mathrm{ft}}$ with Eq.~\eqref{eq:finetune_loss} and update students by gradient descent
    \STATE Update teachers with Eq.~\eqref{eq:emb_ema} if $\lambda^{e^{\star}}_{\mathrm{dream}} > 0$
\ENDFOR
\RETURN $\pi^{e^{\star}}$
\end{algorithmic}
\end{algorithm}

\paragraph{Training samples.}
A training sample is a time step $t$ drawn uniformly from the time steps of the training
trajectories. It contains the observations $\bm{o}^{e}_{t}$, the action chunk of the next
$H = 60$ steps, i.e., 2\,s at 30\,Hz, and the observations of the dreaming modalities
$\mathcal{M}^{e}_{\mathrm{dream}}$ at the offsets $\mathcal{K} = \{20, 40, 60\}$. Chunk steps
and dreaming offsets beyond the end of the trajectory are marked as padding, and action
dimensions that an embodiment does not provide are marked by its action mask; both are excluded
from the losses. For example, end-effector link targets have five fingertip slots per hand, of
which \textsc{G1-Sim-Dex3} uses three, \textsc{xArm7-Dex4} four, and the grippers of
\textsc{LocoMan-Dex2} and \textsc{R1Lite-Gripper} two (thumb and index); torso-pose components
that an embodiment does not control are masked the same way. Non-image observations, actions,
and dreaming sources are standardized per
embodiment and per field with the mean and standard deviation of that embodiment's
demonstrations.

\paragraph{Embodiment-balanced batches.}
In pretraining, every optimizer step draws a mini-batch of $B$ samples from each embodiment in
$\mathcal{E}_{\mathrm{pre}}$, so each embodiment contributes the same number of samples to every optimizer step, regardless
of dataset sizes. The mini-batches are processed one embodiment at a time, and their
gradients are accumulated before a single optimizer step. Because each $\mathcal{L}^{e}$ is
averaged over its own mini-batch, the update follows the sum in Eq.~\eqref{eq:pretrain_loss}.
Fine-tuning and training without pretraining draw mini-batches from the target-task data only.

\paragraph{Losses.}
The action loss follows Eq.~\eqref{eq:act_loss} with $\lambda^{e}_{m} = 1$ for every modality
and is computed on standardized actions. The chunk loss $\ell_{\mathrm{act}}$ uses elementwise
smooth L1 with a threshold of 0.05, which is quadratic for errors below the threshold and
linear above it. For each modality, valid chunk-step and action-dimension entries are pooled
across the embodiment mini-batch before taking the mean. The
\modcolor{tokPU}{dreaming loss} follows Eq.~\eqref{eq:emb_dream_loss} with $\lambda^{e}_{\mathrm{dream}} = 0.3$
and is averaged over all latent tokens at dreaming offsets that are not padded; $\beta = 0.01$ for the
\texttt{Kitchen-Bread} and \texttt{Kitchen-Cleanup} experiments and $\beta = 0.005$ for the
\texttt{Scoop}, \texttt{Insert}, and \texttt{Sort} experiments. For the
\texttt{Kitchen-Bread} and \texttt{Kitchen-Cleanup} experiments, $\mathcal{L}^{e}$ also includes
a penalty $10^{-5} \max\big(0, \|\modcolor{tokPU}{\bm{z}_{j}}\|_2 - 100\big)^2$ on each latent token $\modcolor{tokPU}{\bm{z}_{j}}$
produced by the student \modcolor{tokPU}{embodiment tokenizer} $\modcolor{tokPU}{\phi^{e}_{\mathrm{dream}}}$, averaged over valid
tokens to discourage large token norms. The teachers are exempt from this penalty.

\paragraph{Optimization.}
We use AdamW with weight decay $10^{-4}$ and four parameter groups with separate learning
rates: the tokenizers, which include the modality tokenizers, the \modcolor{tokPU}{embodiment tokenizers}, and any
shared image encoder or tokenizer; the trunk; the action experts; and the \modcolor{tokPU}{dream experts}. All
remaining parameters use the trunk learning rate. The tokenizers, trunk, and action experts
share one learning rate, $10^{-4}$ in pretraining and in training without pretraining and
$5 \times 10^{-5}$ in fine-tuning, while the \modcolor{tokPU}{dream experts} use a much smaller one
(Table~\ref{tab:training_hparams}). Training uses bfloat16 mixed precision without gradient
clipping. Pretraining and training without pretraining use a cosine schedule without warmup:
at step $t$ of a budget of $T$ steps, every parameter group with initial learning rate
$\eta_0$ uses
\begin{equation}
\eta_t = \eta_0 \Big( r + (1 - r)\, \tfrac{1}{2} \big( 1 + \cos(\pi t / T) \big) \Big),
\qquad r = 0.05,
\end{equation}
so each learning rate, including that of the \modcolor{tokPU}{dream experts}, decays to 5\% of its initial
value at the end of training. Fine-tuning uses constant learning rates.

\paragraph{Image tokenizer sharing.}
MED shares the image encoder across embodiments, while each embodiment keeps its own adapter,
i.e., the cross-attention aggregator $g^{e}_{m}$ with its projection and learnable query tokens
(Appendix~\ref{sec:app_architecture}). Encoders are shared per image modality; for example, one
encoder processes the egocentric main-camera images of all embodiments. The shared encoder is
trained on the images of every embodiment, which pools visual data that is highly imbalanced
across embodiments, with human demonstrations making up 49.4\% of all trajectories
(Table~\ref{tab:dataset_statistics}), and lets object features learned from one embodiment's
images, such as those of human demonstrations, be reused by the others. The adapters absorb what
differs between embodiments: camera placement, field of view, image resolution, and how each
embodiment's own body appears in the image.

\paragraph{EMA teachers.}
For each embodiment, the teacher consists of copies of the modality tokenizers of
$\mathcal{M}^{e}_{\mathrm{dream}}$ and of the \modcolor{tokPU}{embodiment tokenizer} $\modcolor{tokPU}{\phi^{e}_{\mathrm{dream}}}$.
The target computation uses the copied feature-encoding components $\bar{f}^{e}_{m}$,
followed by $\modcolor{tokPU}{\bar{\phi}^{e}_{\mathrm{dream}}}$, as in Eq.~\eqref{eq:emb_dream_target}.
When the image encoder or the image tokenizer is shared across embodiments, its teacher copy is
shared as well, while each embodiment keeps a teacher copy of any image module it does not
share, such as its own aggregator. Teachers receive no gradients.
After every optimizer step, their parameters are updated with Eq.~\eqref{eq:emb_ema} using
$\alpha = 0.999$, and their batch-normalization statistics are copied from the students.

\paragraph{Pretraining.}
Each cross-embodiment pretraining run excludes the target embodiment's demonstrations of the
target task, which are used only for fine-tuning. The pretraining-mixture ablations in
Sec.~\ref{sec:experiments} follow the same procedure with fewer embodiments in
$\mathcal{E}_{\mathrm{pre}}$: human-only pretraining uses only human demonstrations, and
human-plus-robot pretraining adds demonstrations of the other robots but no data from the
target embodiment.

\paragraph{Fine-tuning.}
Fine-tuning loads the weights of a pretraining checkpoint, initializes a new optimizer state, and
recomputes the per-field mean and standard deviation used for data normalization from the target-task demonstrations. If the target embodiment is
in $\mathcal{E}_{\mathrm{pre}}$, its tokenizers, action experts, \modcolor{tokPU}{embodiment tokenizer}, dream
expert, and teachers are restored together with the trunk and any shared image encoder or
tokenizer. Otherwise, as in the human-only and human-plus-robot ablations, only the trunk and
the shared image encoder or tokenizer, if one is used, are transferred, and the modules of the
target embodiment are
trained from their initialization. Wrist cameras that are not used during pretraining receive new tokenizers whose
backbones are initialized from ImageNet. Embodiment dreaming remains active during fine-tuning
with the same $\lambda^{e^{\star}}_{\mathrm{dream}}$ and $\beta$.

\paragraph{Training without pretraining and ablations.}
MED without pretraining is trained from scratch on the target-task data with the same
objective, including \modcolor{tokPU}{embodiment dreaming}, and with a private image tokenizer. MXT-FG, the adapted backbone without dreaming,
removes the \modcolor{tokPU}{embodiment tokenizer}, the \modcolor{tokPU}{dream expert}, and the teachers in both pretraining and
fine-tuning, and is otherwise trained with the same recipe as MED.

\begin{table*}[tp]
    \centering
    \caption{
    \textbf{Training hyperparameters.}
    Shared settings apply to all experiments; the remaining blocks list the
    task-specific settings. Pretraining batch sizes count the samples drawn from each
    embodiment per optimizer step. Values in parentheses are for
    \textsc{R1Lite-Gripper}. Parameter counts exclude the EMA teachers and cover
    the trunk together with the modules of every embodiment in pretraining and of the target
    embodiment otherwise. The largest per-embodiment counts are for \texttt{Kitchen-Bread} and
    \texttt{Kitchen-Cleanup}, whose target embodiments add two wrist cameras.
    }
    \label{tab:training_hparams}

    \small
    \setlength{\tabcolsep}{5pt}
    \renewcommand{\arraystretch}{1.15}

    \begin{tabularx}{\textwidth}{
        @{}
        l
        >{\raggedright\arraybackslash}X
        >{\raggedright\arraybackslash}X
        >{\raggedright\arraybackslash}X
        @{}
    }
        \toprule
        & \textbf{Pretraining}
        & \textbf{Fine-tuning}
        & \textbf{No pretraining} \\
        \midrule
        \multicolumn{4}{@{}l}{\textit{Shared settings}} \\
        Optimizer
        & \multicolumn{3}{l}{AdamW, weight decay $10^{-4}$, no gradient clipping} \\
        Precision
        & \multicolumn{3}{l}{bfloat16 mixed precision} \\
        Learning-rate schedule
        & Cosine, no warmup
        & Constant
        & Cosine, no warmup \\
        Tokenizer learning rate
        & $10^{-4}$
        & $5 \times 10^{-5}$
        & $10^{-4}$ \\
        Trunk learning rate
        & $10^{-4}$
        & $5 \times 10^{-5}$
        & $10^{-4}$ \\
        Action-expert learning rate
        & $10^{-4}$
        & $5 \times 10^{-5}$
        & $10^{-4}$ \\
        Action horizon $H$
        & \multicolumn{3}{l}{60 steps (2\,s)} \\
        Action loss
        & \multicolumn{3}{l}{Smooth L1 with threshold 0.05 on standardized actions, $\lambda^{e}_{m} = 1$} \\
        Dreaming offsets $\mathcal{K}$
        & \multicolumn{3}{l}{$\{20, 40, 60\}$ steps} \\
        Dreaming loss weight
        & \multicolumn{3}{l}{$\lambda^{e}_{\mathrm{dream}} = 0.3$} \\
        EMA decay $\alpha$
        & \multicolumn{3}{l}{0.999} \\
        Image backbone
        & \multicolumn{3}{l}{ResNet-18 initialized from ImageNet} \\
        Parameters
        & 120M--185M
        & 50M--90M
        & 50M--90M \\
        \midrule
        \multicolumn{4}{@{}l}{\textit{\texttt{Kitchen-Bread} and \texttt{Kitchen-Cleanup}}} \\
        Batch size
        & 64 per embodiment
        & 128 (72)
        & 128 (72) \\
        Dream-expert learning rate
        & \multicolumn{3}{l}{$5 \times 10^{-6}$ in every stage, with the same schedule} \\
        Dreaming magnitude weight
        & \multicolumn{3}{l}{$\beta = 0.01$} \\
        Latent norm penalty
        & \multicolumn{3}{l}{Bound 100, weight $10^{-5}$, in every stage} \\
        \midrule
        \multicolumn{4}{@{}l}{\textit{\texttt{Insert}}} \\
        Batch size
        & 48 per embodiment
        & 32
        & 32 \\
        Dream-expert learning rate
        & \multicolumn{3}{l}{$10^{-6}$ in every stage, with the same schedule} \\
        Dreaming magnitude weight
        & \multicolumn{3}{l}{$\beta = 0.005$} \\
        Latent norm penalty
        & \multicolumn{3}{l}{None} \\
        \midrule
        \multicolumn{4}{@{}l}{\textit{\texttt{Sort}}} \\
        Batch size
        & 48 per embodiment
        & 32
        & 32 \\
        Dream-expert learning rate
        & \multicolumn{3}{l}{$10^{-6}$ in every stage, with the same schedule} \\
        Dreaming magnitude weight
        & \multicolumn{3}{l}{$\beta = 0.005$} \\
        Latent norm penalty
        & \multicolumn{3}{l}{None} \\
        \midrule
        \multicolumn{4}{@{}l}{\textit{\texttt{Scoop}}} \\
        Batch size
        & 48 per embodiment
        & 32
        & 32 \\
        Dream-expert learning rate
        & \multicolumn{3}{l}{$10^{-6}$ in every stage, with the same schedule} \\
        Dreaming magnitude weight
        & \multicolumn{3}{l}{$\beta = 0.005$} \\
        Latent norm penalty
        & \multicolumn{3}{l}{None} \\
        \bottomrule
    \end{tabularx}
\end{table*}

\subsection{Baseline Implementations}
\label{app:baselines}

ACT is trained from scratch and $\pi_{0.5}$ is fine-tuned from its pretrained checkpoint, both
on demonstrations from the target embodiment and task only. Both are deployed with the
execution loop and controllers of Appendix~\ref{app:deployment}, and
Table~\ref{tab:baseline_hparams} lists their hyperparameters. MED without pretraining and the
MXT-FG ablation are described in Appendix~\ref{sec:app_algorithm}.

\begin{table}[t]
    \centering
    \caption{
    \textbf{Baseline hyperparameters.}
    For parameters and batch size, the first value is for \texttt{Sort},
    \texttt{Insert}, and \texttt{Scoop}, and the second is for \texttt{Kitchen-Bread} and
    \texttt{Kitchen-Cleanup}, where ACT adds a ResNet-18 for each of the two wrist cameras.
    }
    \label{tab:baseline_hparams}

    \small
    \setlength{\tabcolsep}{5pt}
    \renewcommand{\arraystretch}{1.15}

    \begin{tabularx}{\linewidth}{
        @{}
        l
        >{\raggedright\arraybackslash}X
        >{\raggedright\arraybackslash}X
        @{}
    }
        \toprule
        & \textbf{ACT}
        & \textbf{\boldmath$\pi_{0.5}$} \\
        \midrule
        Initialization
        & ResNet-18 from ImageNet; Transformer from scratch
        & Pretrained $\pi_{0.5}$ (\texttt{pi05\_base}) \\
        Architecture
        & Decoder-only, 12 blocks, width 512, 8 heads, dropout 0.1
        & PaliGemma + 300M action expert; LoRA ranks: 16 (language), 32 (action expert) \\
        Parameters
        & 49M; 72M
        & 3.4B \\
        Action horizon
        & 60 steps (2\,s)
        & 60 steps (2\,s) \\
        Training loss
        & L1 $+\ 10^{-3} \times$ feature prediction
        & Flow-matching velocity MSE \\
        Optimizer
        & AdamW, weight decay $10^{-4}$
        & AdamW, weight decay $10^{-10}$ \\
        Learning rate
        & $2 \times 10^{-5}$ ($10^{-5}$ for ResNet-18), constant
        & $2.5 \times 10^{-5} \rightarrow 2.5 \times 10^{-6}$, cosine decay \\
        Precision
        & bfloat16 mixed precision
        & bfloat16/float32 mixed precision \\
        Batch size
        & 32; 128 (72 for \texttt{R1Lite-Gripper})
        & 32; 128 (72 for \texttt{R1Lite-Gripper}) \\
        \bottomrule
    \end{tabularx}
\end{table}

\paragraph{ACT.}
We use the decoder-only variant of ACT~\cite{zhao2023learning} introduced in
HumanPlus~\cite{fu2024humanplus}, which removes the conditional VAE and the encoder--decoder
structure of the original model. A single stack of self-attention blocks processes one
state token, the image tokens, and 60 learned action queries, and a linear head maps
the outputs at the query positions to a chunk of 60 actions. Image tokens are the ResNet-18
feature maps of each camera view, projected to the model width by a $1\times1$ convolution and
combined with 2D sinusoidal positional embeddings. The two halves of the side-by-side stereo
head-camera image are encoded as two views by a shared backbone; in \texttt{Kitchen-Bread} and
\texttt{Kitchen-Cleanup}, the two wrist cameras are added as views with their own backbones.
States and actions are standardized with dataset statistics. The training loss
is an L1 loss on the action chunk, excluding steps padded past the end of an episode, plus an
auxiliary feature-prediction loss: the mean squared error between the outputs at the
image-token positions and the projected backbone features of the frame 60 steps ahead. ACT has
no \modcolor{tokPU}{embodiment tokenizer}, action experts, or \modcolor{tokPU}{dream expert}, and it is not pretrained on other
embodiments.

\paragraph{\boldmath$\pi_{0.5}$.}
We use the OpenPI implementation of $\pi_{0.5}$, initialized from
\texttt{pi05\_base}.
LoRA adapts the language backbone and action expert; the visual
encoder, action projections, and time-conditioning MLPs are
also trainable.
Visual inputs comprise the left stereo view and, for Kitchen tasks, both wrist views, resized with aspect-ratio-preserving padding to $224 \times 224$. We follow the three-camera interface of the pretrained $\pi_{0.5}$ model, assigning the left stereo view to the base-camera slot and the wrist views, when available, to their corresponding wrist-camera slots.
States are quantile-normalized and
discretized into tokens alongside the task instruction.
Actions contain wrist poses with fingertip targets for
\texttt{xArm7-Dex4}, finger commands for \texttt{G1-Dex5}, or
gripper openings for \texttt{R1Lite-Gripper} and
\texttt{LocoMan-Dex2}; \texttt{Z1-Tool} predicts wrist
poses only.
G1 and LocoMan also predict torso commands, and Kitchen policies
include base velocities. Simulated Dex3 hands use one closure
fraction per hand.
Selected pose targets are encoded relative to the current
observation; simulated \texttt{Kitchen-Cleanup} uses SO(3)
rotation vectors for relative orientations.
Actions are quantile-normalized, and state and action tensors
are zero-padded to 32 dimensions.
The model predicts 60-action chunks and is trained with a
flow-matching velocity MSE loss.
Standard fine-tuning follows Table~\ref{tab:baseline_hparams},
with 500 warmup steps and global batch sizes.
At deployment, all $\pi_{0.5}$ policies use real-time chunking
(RTC)~\cite{black2025real}, overlapping prediction of the next action chunk with
execution of the current chunk to reduce discontinuities.
The replanning interval is 45 action steps (1.5\,s), as for all evaluated methods.
The inference server uses 10 Euler integration steps per chunk.
Outputs are unnormalized and converted to embodiment-specific
controller commands, executed at 30\,Hz.

\begin{figure*}[b]
    \centering
    \includegraphics[width=\linewidth]{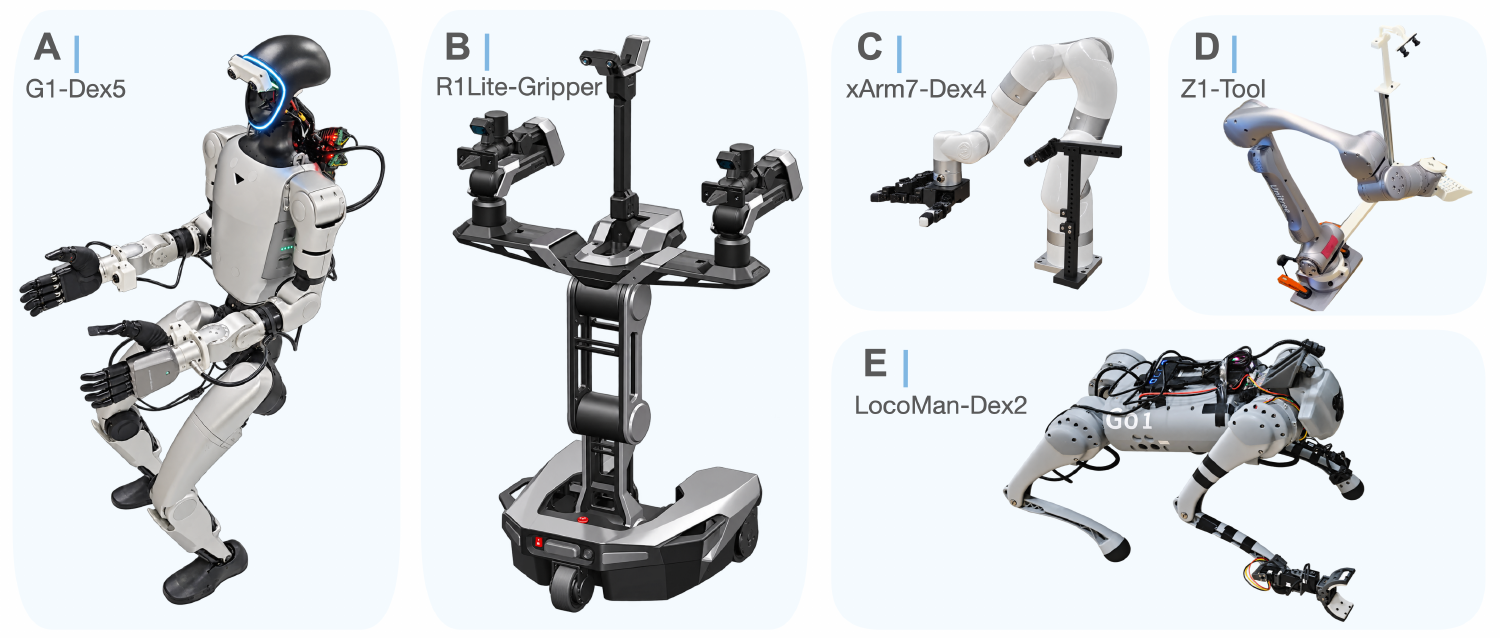}
    \caption{\small
    Physical robot embodiments used in EgoX.
    }
    \label{fig:hardware_setup}
\end{figure*}

\subsection{Embodiment and Hardware Details}
\label{app:hardware}

EgoX is evaluated on five physical robot embodiments spanning substantially
different morphologies, mobility configurations, and manipulation interfaces,
as shown in Fig.~\ref{fig:hardware_setup}.
The platforms include a bimanual humanoid with dexterous hands
(\textsc{G1-Dex5}), a wheeled bimanual mobile manipulator with parallel-jaw
grippers (\textsc{R1Lite-Gripper}), a tabletop manipulator with a dexterous
hand (\textsc{xArm7-Dex4}), a fixed-base arm equipped with a non-actuated tool
(\textsc{Z1-Tool}), and a quadruped with a two-finger gripper
(\textsc{LocoMan-Dex2}).

Across all embodiments, EgoX represents manipulation actions using a shared
set of semantically corresponding geometric action modalities. These include
wrist pose targets and, when available, task-relevant end-effector
link pose targets. For dexterous hands and grippers, these links correspond
to fingertips, while embodiments that do not provide such structure can use
only the applicable subset of action modalities. The predicted geometric
targets are mapped to each embodiment's native control space through
embodiment-specific retargeting and low-level control interfaces.

\paragraph{G1-Dex5.}
\textsc{G1-Dex5} is a Unitree G1 humanoid equipped with two Inspire RH56F1
five-finger dexterous hands. EgoX can predict torso pose, left/right wrist pose targets
together with five fingertip targets for each hand and optional velocity commands for the whole robot. These geometric targets
are retargeted to the corresponding arm and hand control spaces for execution.
For locomotion and whole-body stabilization, we use the low-level whole-body
controller from Humanoid Touch Dream~\cite{niu2026learning}, which is kept
fixed throughout our experiments.

\paragraph{R1Lite-Gripper.}
\textsc{R1Lite-Gripper} is a Galaxea R1 Lite wheeled bimanual mobile
manipulator equipped with parallel-jaw grippers. EgoX predicts left/right
wrist pose targets together with two fingertip targets for each
gripper and optional velocity commands. These geometric targets are mapped to the native arm and gripper
control spaces of the R1 Lite for execution. Mobile-base commands are handled
through the embodiment-specific locomotion interface.

\paragraph{xArm7-Dex4.}
\textsc{xArm7-Dex4} consists of an xArm7 tabletop manipulator equipped with a
four-finger LEAP Hand~\cite{shaw2023leap}. EgoX predicts a wrist pose
target together with four fingertip targets. These geometric targets are
retargeted to the xArm7 arm and LEAP Hand control spaces for execution. Since
the platform has a fixed base, it does not use a locomotion action modality.

\paragraph{Z1-Tool.}
\textsc{Z1-Tool} is a Unitree Z1 tabletop manipulator equipped with a fixed,
non-actuated scoop. For this embodiment, EgoX predicts only the wrist
pose target, which is executed by the underlying arm controller. Since the
scoop is rigidly attached to the arm and has no actuated joints, this
embodiment does not predict a separate end-effector keypoint or actuation modality.

\paragraph{LocoMan-Dex2.}
\textsc{LocoMan-Dex2} is a quadrupedal manipulation platform with two-finger
grippers mounted on its front legs~\cite{lin2024locoman}. We evaluate the
platform in a unimanual configuration. EgoX predicts a wrist pose
target together with two fingertip targets for the active gripper. These
geometric targets are mapped to the native control space of the active
manipulation limb for execution.
\begin{table*}[t]
    \centering
    \caption{
    \textbf{Embodiment-specific interfaces used in EgoX.}
    EgoX represents manipulation actions using semantically corresponding
    geometric modalities while allowing each embodiment to expose only the
    modalities applicable to its morphology.
    EEF denotes end effector.
    }
    \label{tab:embodiment_interfaces}

    \small
    \setlength{\tabcolsep}{5pt}
    \renewcommand{\arraystretch}{1.15}

    \begin{tabular*}{\textwidth}{
        @{\extracolsep{\fill}}
        l
        l
        c
        c
        c
        c
        c
        c
        @{}
    }
        \toprule
        \textbf{Embodiment}
        & \textbf{Side}
        & \shortstack{\textbf{Robot}\\\textbf{Joint Dim.}}
        & \shortstack{\textbf{EEF}\\\textbf{Joint Dim.}}
        & \shortstack{\textbf{Torso}\\\textbf{Action}}
        & \shortstack{\textbf{Wrist Pose}\\\textbf{Action}}
        & \shortstack{\textbf{Fingertip}\\\textbf{Targets}} \\
        \midrule

        \textsc{G1-Dex5}
        & Bimanual
        & 29
        & $6+6$
        & Yes
        & L/R
        & 5 / hand \\

        \textsc{R1Lite-Gripper}
        & Bimanual
        & 15
        & $1+1$
        & No
        & L/R
        & 2 / gripper \\

        \textsc{xArm7-Dex4}
        & Right
        & 7
        & 16
        & No
        & Right
        & 4 \\

        \textsc{LocoMan-Dex2}
        & Right
        & 18
        & 1
        & Yes
        & Right
        & 2 \\

        \textsc{Z1-Tool}
        & Right
        & 6
        & --
        & No
        & Right
        & -- \\

        \bottomrule
    \end{tabular*}

    \vspace{3pt}
    \begin{minipage}{\textwidth}
    \footnotesize
    ``Robot Joint Dim.'' and ``EEF Joint Dim.'' denote the native
    proprioceptive dimensions used by each embodiment.
    EgoX predicts geometric wrist pose and, when applicable, fingertip targets,
    which are subsequently mapped to each embodiment's native control space.
    \end{minipage}
\end{table*}

\subsection{Unified Egocentric Frame System}
\label{app:frames}

All frames in EgoX are right-handed, with the $x$-axis pointing forward, the
$y$-axis to the left, and the $z$-axis upward.
Table~\ref{tab:frames} summarizes the frames of each embodiment. The
root frame $\mathcal{F}_{\text{root}}$ is the reference for the torso pose
and the velocity commands. It is the gravity-aligned projection of a
reference body part onto the ground that keeps the part's heading: the head
at the start of the episode for human data, and the feet, trunk, chassis, or
arm base for robots. Except for
the fixed-base arms, the wrist poses in $\mathcal{F}_{\text{root}}$ are also
provided as observations. The egocentric frame
$\mathcal{F}_{\text{ego}}$ (Sec.~\ref{sec:method_data}) is rigidly attached to
the body that carries the main camera. Wrist poses and wrist targets are
expressed relative to it,
\begin{equation}
    {}^{\text{ego}}\bm{T}_{\text{wrist}}
    = \big({}^{\text{root}}\bm{T}_{\text{ego}}\big)^{-1}\,
      {}^{\text{root}}\bm{T}_{\text{wrist}},
\end{equation}
where ${}^{A}\bm{T}_{B}\in SE(3)$ is the pose of frame $B$ in frame $A$, and
end-effector-link keypoints are expressed in the local frame of each end
effector. Poses are stored as a position and roll, pitch, and yaw angles in
$Z$--$Y$--$X$ order. A fixed rotation brings each robot's native wrist frame
to the EgoX wrist convention, $\bm{R}_{\text{wrist}} =
\bm{R}_{\text{native}}\bm{M}^{\top}$, where the rows of $\bm{M}$ give the
native axes that become the EgoX wrist $x$-, $y$-, and $z$-axes
(Table~\ref{tab:frames}). Torso-pose components that an embodiment does not
use are masked (Appendix~\ref{sec:app_algorithm}).

\begin{table*}[t]
    \centering
    \caption{
    \textbf{Root and egocentric frames of each embodiment.}
    All frames use $x$ forward, $y$ left, and $z$ up.
    ``Torso pose'' lists the torso-pose components marked valid in the
    recorded data (rpy: roll, pitch, and yaw).
    ``Wrist axes'' lists the native wrist axes that become the EgoX wrist
    $(x, y, z)$ axes; for the human, the tracked wrist frame defines the
    convention.
    }
    \label{tab:frames}

    \small
    \setlength{\tabcolsep}{4pt}
    \renewcommand{\arraystretch}{1.15}

    \begin{tabularx}{\textwidth}{@{} l
        >{\raggedright\arraybackslash}X
        >{\raggedright\arraybackslash}X
        l l @{}}
        \toprule
        \textbf{Embodiment}
        & \textbf{Root frame} $\mathcal{F}_{\text{root}}$
        & \textbf{Egocentric frame} $\mathcal{F}_{\text{ego}}$
        & \textbf{Torso pose}
        & \textbf{Wrist axes} \\
        \midrule

        \textsc{Human-Dex5}
        & Ground point below the head at episode start
        & Tracked head, carrying the stereo camera
        & $x$, $y$, $z$, rpy
        & Tracked wrist \\

        \textsc{G1-Dex5}
        & Stance frame at the feet, from leg kinematics
        & Torso link, origin shifted by $(0.154, 0, 0.406)$\,m
        & $z$, rpy
        & \begin{tabular}[t]{@{}l@{}}L: $(x, -z, y)$\\ R: $(x, z, -y)$\end{tabular} \\

        \textsc{G1-Sim-Dex3}
        & \multicolumn{4}{l}{Same as \textsc{G1-Dex5}} \\

        \textsc{R1Lite-Gripper}
        & Wheeled chassis
        & Floating base (top torso link)
        & $z$, pitch
        & \begin{tabular}[t]{@{}l@{}}L: $(x, -z, y)$\\ R: $(x, z, -y)$\end{tabular} \\

        \textsc{LocoMan-Dex2}
        & Ground projection of the trunk at mode entry
        & Trunk
        & $z$, rpy
        & $(-z, x, -y)$ \\

        \textsc{xArm7-Dex4}
        & Arm base
        & Arm base (static camera)
        & --
        & $(x, -y, -z)$ \\

        \textsc{Z1-Tool}
        & Arm base rotated by $-90^\circ$ about $z$
        & Same as root frame (static camera)
        & --
        & $(x, y, z)$ \\

        \bottomrule
    \end{tabularx}
\end{table*}

\paragraph{Human-Dex5.}
The headset reports head and wrist poses in its own gravity-aligned tracking
frame, whose $y$-axis points up; a fixed rotation maps them to the convention
above. When the operator starts an episode (Appendix~\ref{app:dataset}), this
frame is rotated about $z$ by the operator's head yaw and shifted to the head
position, giving a frame with its origin at the head, its $x$-axis along the
horizontal heading of the head, and its $z$-axis opposite to gravity, all
taken at the start of the episode. $\mathcal{F}_{\text{root}}$ is a pseudo
root frame, the projection of this frame onto the ground
(Fig.~\ref{fig:frames}).
$\mathcal{F}_{\text{ego}}$ is the tracked head frame, which carries the
headset-mounted stereo camera and follows the full head orientation,
including pitch and roll. No head-to-torso offset is applied, so the torso
modality of human data is the head pose in $\mathcal{F}_{\text{root}}$, with
its height measured from the starting head height. The
25 tracked hand keypoints are expressed in the tracked wrist frame, and the
fingertips are taken from them.

\paragraph{G1-Dex5.}
The controller computes the robot state from the joint encoders by forward
kinematics, assuming that both feet are flat on the ground.
$\mathcal{F}_{\text{root}}$ is the resulting stance frame, recomputed at
every control step: its origin is the midpoint of the two ankle-roll links,
its $z$-axis is the mean of the two foot $z$-axes, and its $x$-axis is the
mean forward direction of the feet, made orthogonal to $z$.
$\mathcal{F}_{\text{ego}}$ is a virtual head frame rigidly attached to the
torso link, which carries the head and its camera. It has the orientation of
the torso link, and its origin lies at $(0.154, 0, 0.406)$\,m in torso-link
coordinates, i.e., $(0.15, 0, 0.45)$\,m from the pelvis when the waist joints
are at zero. The torso observation is the torso-link pose in
$\mathcal{F}_{\text{root}}$. The torso target is the operator's head pose in
the human $\mathcal{F}_{\text{root}}$: the whole-body controller tracks its
orientation as the torso orientation and its height change $\Delta z$ since
the reset as a torso height of $0.75\,\text{m} + \Delta z$. Each wrist frame
lies 0.05\,m beyond the wrist-yaw joint along the joint's $x$-axis. For arm
inverse kinematics, wrist targets in $\mathcal{F}_{\text{ego}}$ are shifted
by the same head offset into the pelvis frame.

\paragraph{G1-Sim-Dex3.}
The simulated G1 with three-finger Dex3 hands follows the same conventions as
\textsc{G1-Dex5}.

\paragraph{R1Lite-Gripper.}
The vendor motion-control stack reports the pose of the floating base, the
top torso link that carries the stereo head camera and both arms, in the
frame of the wheeled chassis, and the end-effector (wrist) poses relative to the
floating base. $\mathcal{F}_{\text{root}}$ is therefore the chassis frame,
and $\mathcal{F}_{\text{ego}}$ is the floating-base frame. Only the height
and pitch of the torso pose are used; the policy predicts no torso target
(Table~\ref{tab:embodiment_interfaces}), and the chassis is driven by
velocity commands. The wrist frames are the vendor end-effector frames,
remapped with the same side-specific rotations as \textsc{G1-Dex5}.

\paragraph{LocoMan-Dex2.}
The robot is operated in its single-gripper manipulation mode, in which the
onboard state estimator expresses the trunk pose relative to the trunk frame
at the moment the mode is entered; $\mathcal{F}_{\text{root}}$ is the
projection of that frame onto the ground. Afterwards, the trunk pose is the rigid
transform that aligns the current stance-foot positions, obtained from leg
kinematics, with their positions at that moment, so
$\mathcal{F}_{\text{root}}$ stays fixed to the stance feet. The gripper pose
follows from the trunk pose and the forward kinematics of the manipulating
leg and its gripper. $\mathcal{F}_{\text{ego}}$ is the trunk frame, which
needs no rotation remap in the unimanual configuration. The height and
orientation of the torso pose are used, both measured relative to the trunk
frame at mode entry.

\paragraph{xArm7-Dex4.}
The arm has a fixed base and a static main camera, so
$\mathcal{F}_{\text{root}}$ and $\mathcal{F}_{\text{ego}}$ coincide with the
xArm7 base frame, whose $x$-axis points forward from the base and whose
$z$-axis points up. The torso pose is identically zero and is not used. The
wrist frame is the joint-7 frame without offset, remapped as in
Table~\ref{tab:frames} for the default reset configuration.

\paragraph{Z1-Tool.}
The Z1 also has a fixed base and a static main camera, so
$\mathcal{F}_{\text{root}}$ and $\mathcal{F}_{\text{ego}}$ coincide. Both are
the Z1 base frame rotated by $-90^\circ$ about $z$: the unified $x$-axis is
the native $-y$-axis, the unified $y$-axis is the native $x$-axis, and the
$z$-axis is unchanged. Poses reported by the arm are rotated into this frame,
and targets are rotated back before inverse kinematics. The wrist frame is
the joint-6 frame without offset or remap. The scoop keypoint lies 0.2\,m
beyond the joint-6 frame along its $x$-axis.

\paragraph{From operator motion to robot targets.}
During teleoperation, the operator's head and wrist poses are first expressed
in the human frames above. For \textsc{G1-Dex5}, the wrist poses relative to
the head are sent as wrist targets in the robot's $\mathcal{F}_{\text{ego}}$.
For the other physical robots, the operator's head and wrist displacements
and rotations since the reset are applied to the robot's torso and
wrist poses at the reset, expressed in the robot's
$\mathcal{F}_{\text{root}}$. For \textsc{Z1-Tool}, these displacements and
rotations are scaled by 0.7 and 0.5, respectively.

\subsection{Robot Control and Deployment}
\label{app:deployment}

\begin{figure*}[t]
    \centering
    \includegraphics[width=\linewidth]{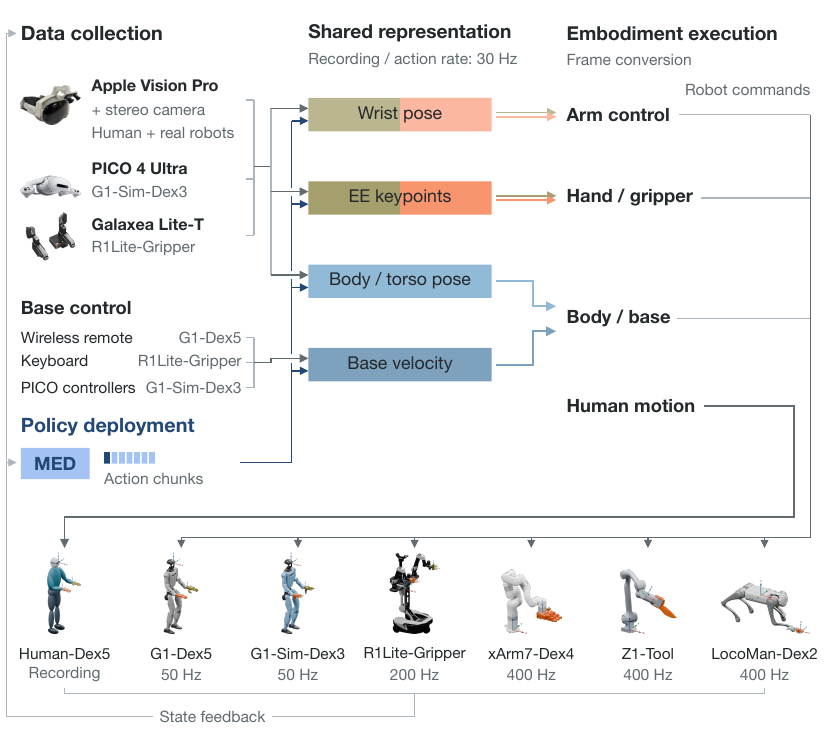}
    \caption{\small
    \textbf{Shared representation and embodiment-specific execution.}
    \textsc{Human-Dex5} motion is recorded using an Apple Vision Pro
    with a mounted stereo camera. Human recordings and robot
    trajectories share the pose/action representation. Robot teleoperation and
    policy rollout use embodiment-specific controllers. Colors follow the action
    experts in Fig.~\ref{fig:architecture} (\modcolor{tokLE}{olive}: left, \modcolor{tokRE}{orange}: right).
    Action modalities vary by embodiment; frequencies below each robot indicate
    its low-level control rate.
    }
    \label{fig:control_pipeline}
\end{figure*}

\paragraph{Shared control stack.}
On the physical robots, teleoperation and policy deployment share the same
control stack (Fig.~\ref{fig:control_pipeline}). A high-level process, either
the teleoperation interface or the policy, produces geometric targets in the
unified egocentric frame system: wrist pose targets in
$\mathcal{F}_{\text{ego}}$, end-effector link pose targets in the wrist frames, and, when
applicable, torso pose targets in $\mathcal{F}_{\text{root}}$ and velocity
commands. Fixed, embodiment-specific
transforms map these targets to the robot's base frame and end-effector
convention, and the targets are sent over CycloneDDS to an
embodiment-specific controller that converts them into joint commands. The
controller publishes the robot state back, which is recorded during
teleoperation and used as policy input during deployment. Because
demonstrations are collected through the same controllers that later execute
the policy, recorded actions and policy outputs are interpreted in the same
way. Table~\ref{tab:deployment} summarizes the controller of each embodiment.

\paragraph{Policy execution.}
At deployment, all methods send actions at 30\,Hz, matching the recording rate,
and use the same embodiment-specific controllers. All methods predict chunks
of 60 actions (2\,s) and replan every 45 action steps (1.5\,s) using updated
observations, following a receding-horizon execution scheme.
MED and ACT do not use temporal ensembling. The $\pi_{0.5}$ policies use RTC,
overlapping prediction of the next chunk with execution of the current chunk
(Appendix~\ref{app:baselines}).
An operator can pause the policy or reset the robot from a keyboard at any time.

\paragraph{Arm control.}
For \textsc{G1-Dex5}, \textsc{xArm7-Dex4}, and \textsc{Z1-Tool}, wrist
pose targets are converted into arm joint targets by inverse kinematics (IK),
solved at every control step with Pinocchio~\cite{carpentier2019pinocchio}
and CasADi~\cite{andersson2019casadi} using IPOPT~\cite{wachter2006ipopt}
with at most 50 iterations. The objective is a weighted sum of the
translation error, the rotation error, a joint regularizer, and a smoothness
term toward the previous solution, with weights $(50, 1, 0.02, 0.1)$ for
\textsc{G1-Dex5} and $(30, 1, 0.02, 0.2)$ for \textsc{xArm7-Dex4} and
\textsc{Z1-Tool}. For \textsc{G1-Dex5}, only the 14 arm joints are optimized
and all other joints are locked. Each solution is then smoothed by a weighted
moving average over the four most recent solutions, with weights
$(0.4, 0.3, 0.2, 0.1)$ from newest to oldest. \textsc{R1Lite-Gripper} uses
RelaxedIK~\cite{rakita2018relaxedik} for its two 6-DoF arms, and
\textsc{LocoMan-Dex2} reuses the whole-body controller and finite-state
machine of Human2LocoMan~\cite{niu2025human2locoman}, which convert the
end-effector target of the active manipulator into joint commands.

\paragraph{Hand and gripper control.}
For the Inspire hands of \textsc{G1-Dex5} and the LEAP Hand of
\textsc{xArm7-Dex4}, fingertip targets expressed in the wrist frame are
converted into hand joint commands by DexPilot
retargeting~\cite{handa2020dexpilot}, using an implementation adapted from
dex-retargeting~\cite{qin2023anyteleop}.
At every control step, the hand joint angles $\bm{q}$ are obtained by solving
\begin{equation}
\min_{\bm{q}_{\min} \le \bm{q} \le \bm{q}_{\max}}\;
\frac{1}{N} \sum_{i=1}^{N} w_i\,
\ell_{\text{smooth-L1}}\big(\|\bm{v}_i(\bm{q}) - \hat{\bm{v}}_i\|_2\big)
+ \lambda_{\mathrm{s}}\, \|\bm{q} - \bm{q}_{\mathrm{prev}}\|_2^2
+ \gamma\, \|\bm{q}\|_2^2,
\label{eq:hand_retargeting}
\end{equation}
where $\bm{v}_i(\bm{q})$ ranges over the $N$ wrist-to-fingertip and
fingertip-to-fingertip vectors of the robot hand computed by forward
kinematics, $\hat{\bm{v}}_i$ is the corresponding target vector,
$\ell_{\text{smooth-L1}}$ is the smooth L1 loss with a threshold of 3\,cm, and
$\bm{q}_{\mathrm{prev}}$ is the previous solution. Target vectors are the human
vectors scaled by 1.2 for the Inspire hands and 1.6 for the LEAP Hand. As in
DexPilot, a thumb--finger pair closer than 3\,cm is treated as a pinch until
the distance exceeds 5\,cm. During a pinch, its target vector is shortened to
0.1\,mm so that the fingertips make contact, and two fingers that pinch the
thumb together while within 3\,cm of each other are kept 3\,cm apart. These
projected vectors are weighted by $w_i = 200$ and $w_i = 400$, respectively,
the remaining fingertip-to-fingertip vectors by $w_i = 1$, and the
wrist-to-fingertip vectors by $w_i = 15$ for the Inspire hands and $w_i = 10$
for the LEAP Hand. The second term, with $\lambda_{\mathrm{s}} = 4 \times
10^{-3}$, keeps consecutive solutions close. The last term is not part of
dex-retargeting; following DexPilot, we add it to regularize the joint angles
toward zero, which corresponds to an open hand. This reduces redundancy in the
solution and keeps the hand out of poor local minima that are difficult to
recover from, such as fingers embedded in the palm. We use $\gamma = 1.5 \times
10^{-2}$, whereas DexPilot uses $\gamma = 2.5 \times 10^{-3}$ for the Allegro
hand. The problem is solved with SLSQP, warm-started from the previous
solution, and the resulting joint angles are smoothed by a first-order low-pass
filter ($\alpha = 0.2$) before being sent to the hand.

For the grippers of \textsc{R1Lite-Gripper} and \textsc{LocoMan-Dex2}, the
controller computes the distance between the two fingertip targets and maps it
linearly to a gripper command, using the same map as teleoperation. On
\textsc{LocoMan-Dex2}, distances from 2 to 15\,cm map to gripper angles from
0.01 to 1.8\,rad. On \textsc{R1Lite-Gripper}, distances from 0.8 to 15\,cm map
to the full gripper range; the command is low-pass filtered and then
binarized, so the gripper opens fully when the distance exceeds about 5\,cm
and closes otherwise.

\paragraph{Body and base control.}
On \textsc{G1-Dex5}, the predicted torso pose, consisting of torso height and
orientation, and the velocity command, consisting of forward, lateral, and yaw
velocities, are passed to the lower-body controller of Humanoid Touch
Dream~\cite{niu2026learning}. This learned policy runs at 50\,Hz and outputs
targets for the 12 leg joints and 3 waist joints. On \textsc{R1Lite-Gripper},
the predicted velocity command drives the mobile chassis and is set to zero
whenever the most recent command is older than 0.4\,s.

\begin{table*}[t]
    \centering
    \caption{
    \textbf{Low-level control used for teleoperation and deployment.}
    Every embodiment receives targets at 30\,Hz; the listed rate is the
    frequency of the embodiment controller.
    }
    \label{tab:deployment}

    \small
    \setlength{\tabcolsep}{5pt}
    \renewcommand{\arraystretch}{1.15}

    \begin{tabularx}{\textwidth}{
        @{}
        l
        >{\raggedright\arraybackslash}X
        >{\raggedright\arraybackslash}X
        >{\raggedright\arraybackslash}X
        r
        @{}
    }
        \toprule
        \textbf{Embodiment}
        & \textbf{Arm}
        & \textbf{Hand / gripper}
        & \textbf{Body / base}
        & \textbf{Rate} \\
        \midrule

        \textsc{G1-Dex5}
        & IK, 14 arm joints
        & DexPilot retargeting
        & Lower-body policy
        & 50\,Hz \\

        \textsc{R1Lite-Gripper}
        & RelaxedIK, $2\times6$ joints
        & Fingertip distance
        & Chassis velocity
        & 200\,Hz \\

        \textsc{xArm7-Dex4}
        & IK, 7 joints
        & DexPilot retargeting
        & --
        & 400\,Hz \\

        \textsc{LocoMan-Dex2}
        & Whole-body controller
        & Fingertip distance
        & Whole-body controller
        & 400\,Hz \\

        \textsc{Z1-Tool}
        & IK, 6 joints
        & --
        & --
        & 400\,Hz \\

        \bottomrule
    \end{tabularx}
\end{table*}

\subsection{Dataset and Data Collection}
\label{app:dataset}

\begin{figure*}[t]
    \centering
    \includegraphics[width=\textwidth]
    {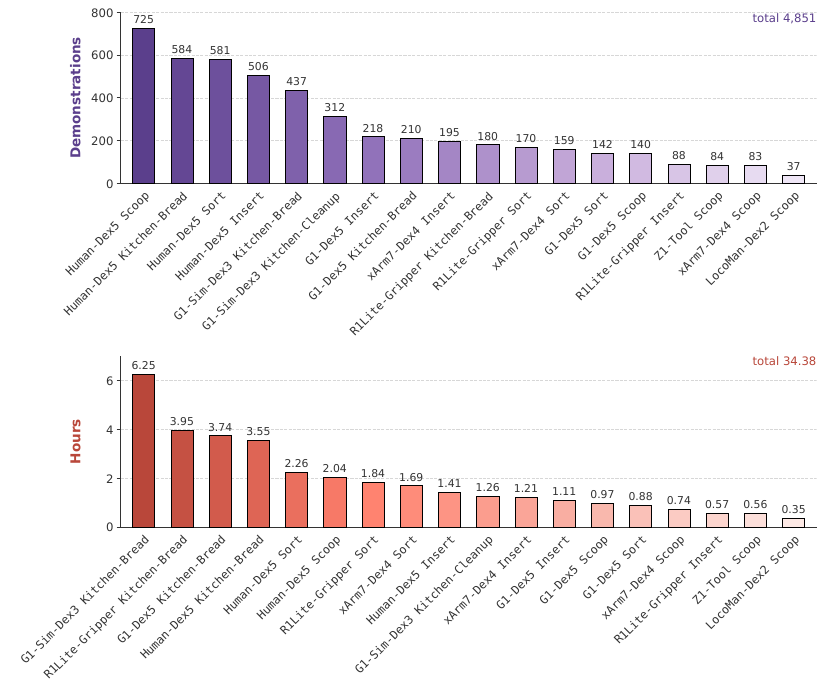}
    \caption{\small
    \textbf{Composition of the EgoX dataset across embodiments and tasks.}
    Number of demonstrations (top) and hours (bottom) for each
    task--embodiment pair, sorted from high to low in each panel.
    }
    \label{fig:dataset_distribution}
\end{figure*}

\paragraph{Dataset overview.}
EgoX is trained on a heterogeneous collection of egocentric manipulation
demonstrations spanning human, physical-robot, and simulated embodiments.
In total, the dataset contains 4,851 trajectories corresponding to
34.38 hours of interaction. Table~\ref{tab:dataset_statistics} summarizes
the dataset composition for each embodiment, while
Fig.~\ref{fig:dataset_distribution} visualizes the task distribution across
embodiments.

\begin{table*}[h]
    \centering
    \caption{
    \textbf{Dataset statistics.}
    Number of training and validation trajectories and total interaction
    duration for each embodiment used in EgoX.
    }
    \label{tab:dataset_statistics}

    \small
    \setlength{\tabcolsep}{6pt}
    \renewcommand{\arraystretch}{1.12}

    \begin{tabularx}{\textwidth}{
        @{}
        l
        c
        >{\raggedright\arraybackslash}X
        r
        r
        r
        r
        @{}
    }
        \toprule
        \textbf{Embodiment}
        & \textbf{Domain}
        & \textbf{Tasks}
        & \textbf{Train}
        & \textbf{Val.}
        & \textbf{Total}
        & \textbf{Hours} \\
        \midrule

        \textsc{Human-Dex5}
        & Human
        & Insert, Kitchen-Bread, Scoop, Sort
        & 2,198 & 198 & 2,396 & 9.26 \\

        \textsc{G1-Dex5}
        & Real
        & Insert, Kitchen-Bread, Scoop, Sort
        & 622 & 88 & 710 & 6.70 \\

        \textsc{xArm7-Dex4}
        & Real
        & Insert, Scoop, Sort
        & 398 & 39 & 437 & 3.64 \\

        \textsc{G1-Sim-Dex3}
        & Sim
        & Kitchen-Bread, Kitchen-Cleanup
        & 691 & 58 & 749 & 7.51 \\

        \textsc{LocoMan-Dex2}
        & Real
        & Scoop
        & 34 & 3 & 37 & 0.35 \\

        \textsc{R1Lite-Gripper}
        & Real
        & Insert, Kitchen-Bread, Sort
        & 398 & 40 & 438 & 6.36 \\

        \textsc{Z1-Tool}
        & Real
        & Scoop
        & 77 & 7 & 84 & 0.56 \\

        \midrule
        \textbf{Total}
        & --
        & --
        & \textbf{4,418}
        & \textbf{433}
        & \textbf{4,851}
        & \textbf{34.38} \\

        \bottomrule
    \end{tabularx}
\end{table*}

\paragraph{Data composition.}
The dataset is deliberately heterogeneous in both embodiment and task
composition. Human demonstrations constitute the largest source, with
2,396 trajectories (49.4\% of the full dataset), while demonstrations
collected on physical robots contribute 1,706 trajectories (35.2\%).
The remaining 749 trajectories (15.4\%) are collected with the simulated
G1 embodiment. Among the robot embodiments, \textsc{G1-Dex5} and
\textsc{G1-Sim-Dex3} provide the largest individual datasets, with 710
and 749 trajectories, respectively, while the remaining platforms provide
smaller but morphologically diverse sources of manipulation experience.

Importantly, the dataset is not constructed by replicating the same task set
across every embodiment. Instead, different embodiments contribute different
subsets of tasks according to their morphology and manipulation capabilities.
Cross-embodiment overlap therefore arises through semantically corresponding
observation and action modalities rather than requiring identical tasks or
actuator-level action spaces across all platforms.

\paragraph{Data collection.}
As in Human2LocoMan~\cite{niu2025human2locoman}, both human and
real-robot data are collected with an Apple Vision Pro that captures the
operator's head and hand motions, while simulated data are collected by
teleoperation in simulation with PICO XR devices
(Appendix~\ref{app:simulation}). The headset tracks the head pose and 25
keypoints per hand, and all streams are recorded at 30\,Hz. Operators start
and stop each episode with left-hand pinch gestures, thumb to index finger to
start and thumb to middle finger to stop and reset, so no additional input
device is needed. Each episode is stored as an HDF5 file with the
available images, poses, proprioception, and actions.

\paragraph{Human demonstrations.}
For human data, a stereo camera is rigidly mounted on the headset and
records egocentric images as
side-by-side stereo frames of $1280\times480$ pixels. The operator performs
the task with their own hands while seeing the scene through the headset's
see-through mode~\cite{park2024avp}. The tracked wrist poses are expressed in
the egocentric frame $\mathcal{F}_{\text{ego}}$ (Sec.~\ref{sec:method_data}), and the five
fingertips of each hand, expressed in the wrist frame, serve as the
end-effector link keypoints.

\paragraph{Robot demonstrations.}
Physical robots are teleoperated with the same headset. The operator's wrist
poses and hand keypoints
are converted into targets in the unified egocentric frame system and streamed at
30\,Hz to the robot, where the control stack of Appendix~\ref{app:deployment}
executes them. For grippers, the thumb--index pinch distance sets the gripper
opening, and for \textsc{G1-Dex5}, the operator's head orientation and
changes in head height set the torso orientation and height. Base velocity commands are issued with a separate input device, the
wireless remote of \textsc{G1-Dex5} or a keyboard for the
\textsc{R1Lite-Gripper} chassis, and are recorded from the robot state. The
operator watches the robot's stereo head-camera stream in the headset through
an Open-TeleVision~\cite{cheng2024opentelevision} interface. Each recorded step
stores the commanded targets as actions, together with the robot's egocentric
images and the poses and proprioception returned by the controller.

\subsection{Task and Evaluation Protocol}
\label{app:evaluation_protocol}

\begin{figure}[!htb]
    \centering
    \includegraphics[width=\linewidth]{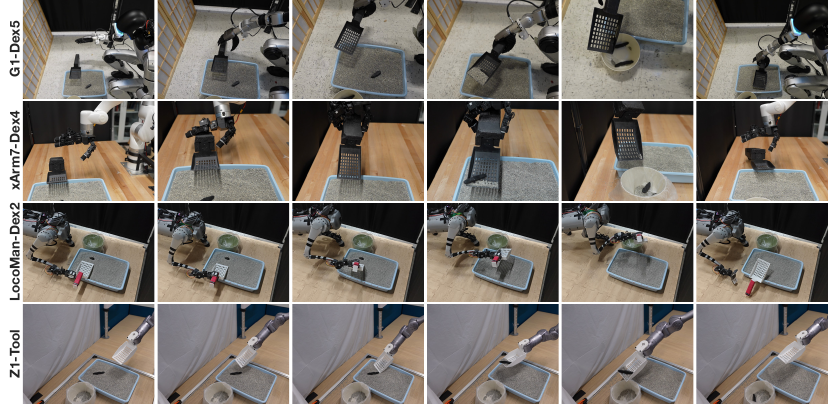}
    \caption{\small
    \textbf{\texttt{Scoop} across four embodiments.}
    Each row progresses from left to right through grasping or positioning the scoop, collecting
    waste, emptying it into the disposal bin, and returning the scoop.
    \textsc{Z1-Tool} uses a fixed scoop and does not perform a grasp.
    The \textsc{LocoMan-Dex2} example shows an OOD trial.
    }
    \label{fig:task_scoop}
\end{figure}

\begin{figure}[!htb]
    \centering
    \includegraphics[width=\linewidth]{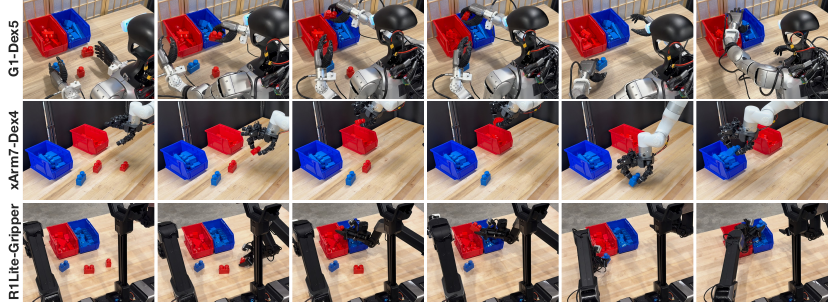}
    \caption{\small
    \textbf{\texttt{Sort} across three embodiments.}
    Each row shows the robot grasping blocks and placing them into the
    color-matched bins, with time progressing from left to right.
    }
    \label{fig:task_sort}
\end{figure}

\begin{figure}[!htb]
    \centering
    \includegraphics[width=\linewidth]{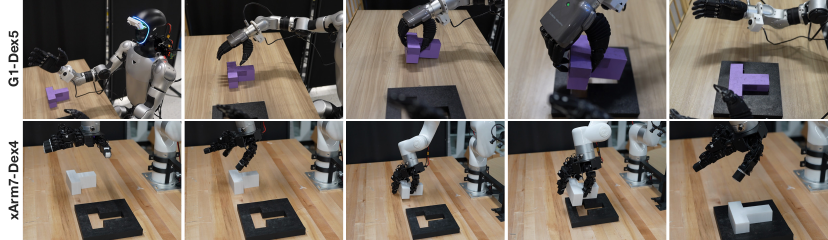}
    \caption{\small
    \textbf{\texttt{Insert} across two embodiments.}
    Each row shows the robot approaching, grasping, aligning, and inserting
    the T-shaped block, with time progressing from left to right.
    }
    \label{fig:task_insert}
\end{figure}

\begin{figure}[!htb]
    \centering
    \includegraphics[width=\linewidth]{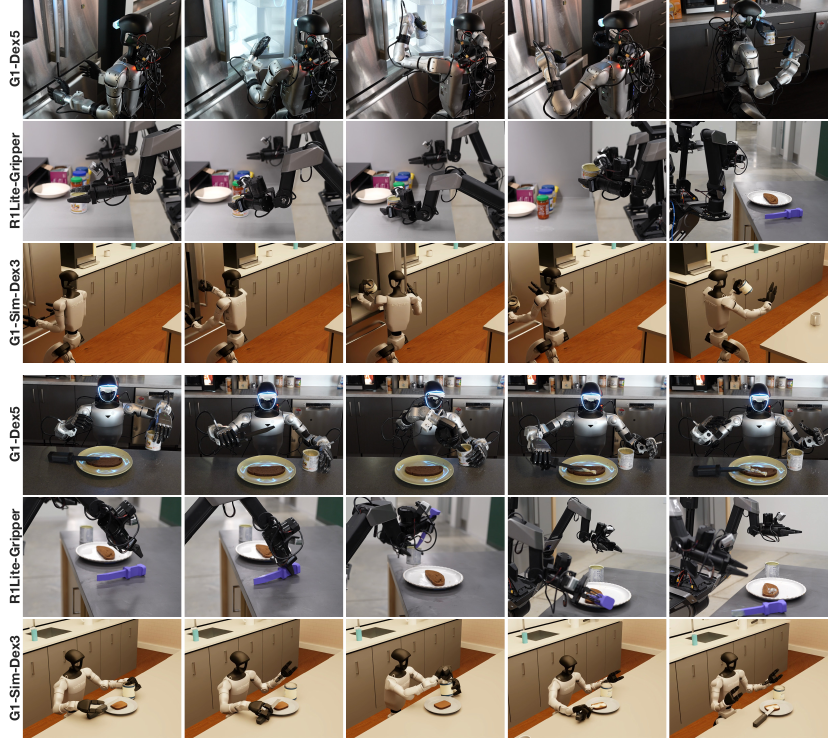}
    \caption{\small
    \textbf{\texttt{Kitchen-Bread} across three embodiments.}
    The upper three rows show container pickup and transport; the lower
    three rows continue the same rollouts through container placement,
    tool pickup, dipping, spreading, and tool return. Time progresses from
    left to right within each row. \textsc{R1Lite-Gripper} retrieves the
    container from a counter rather than a refrigerator.
    }
    \label{fig:task_kitchen_bread}
\end{figure}

\begin{figure}[htbp]
    \centering
    \includegraphics[width=\linewidth]{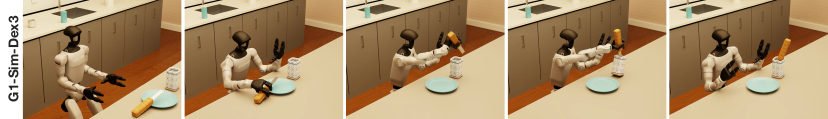}
    \caption{\small
    \textbf{\texttt{Kitchen-Cleanup} on \textsc{G1-Sim-Dex3}.}
    From left to right, the robot approaches the spreader, grasps and lifts
    it, and places it into the container.
    }
    \label{fig:task_kitchen_cleanup}
\end{figure}

We evaluate 13 task--embodiment pairs: \texttt{Scoop} on \textsc{G1-Dex5},
\textsc{xArm7-Dex4}, \textsc{LocoMan-Dex2}, and \textsc{Z1-Tool};
\texttt{Sort} on \textsc{G1-Dex5}, \textsc{xArm7-Dex4}, and
\textsc{R1Lite-Gripper}; \texttt{Insert} on \textsc{G1-Dex5} and
\textsc{xArm7-Dex4}; \texttt{Kitchen-Bread} on \textsc{G1-Dex5},
\textsc{R1Lite-Gripper}, and \textsc{G1-Sim-Dex3}; and
\texttt{Kitchen-Cleanup} on \textsc{G1-Sim-Dex3}.
Figs.~\ref{fig:task_scoop}--\ref{fig:task_kitchen_cleanup} show example rollouts of these pairs, grouped by task.

\begin{figure}[!htbp]
    \centering
    \includegraphics[width=\linewidth]{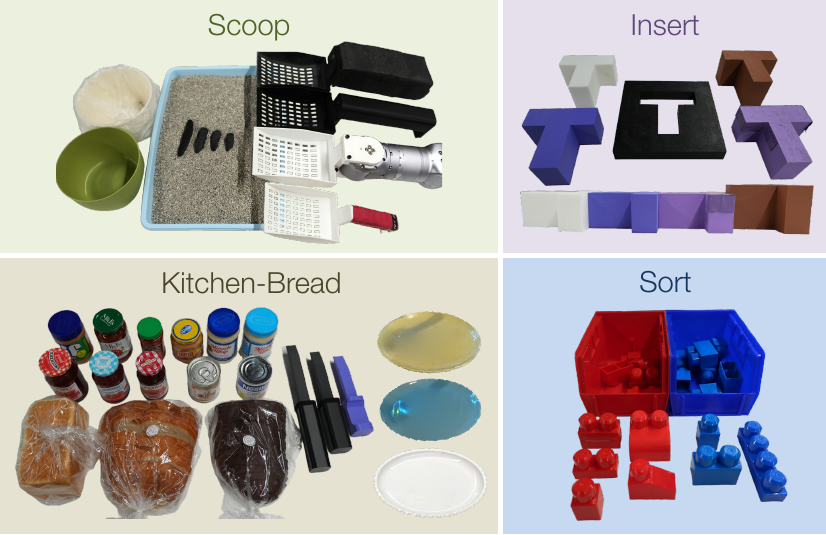}
    \caption{\small
    \textbf{Real objects of the tasks.}
    Each panel shows the objects of one task. The lower \texttt{Insert} photo
    shows the blocks from the side to compare their sizes. Which objects appear
    in ID and which in OOD trials differs between embodiments and is given in
    the description of each task.
    }
    \label{fig:objects_real}

    \vspace{\floatsep}
    \includegraphics[width=\linewidth]{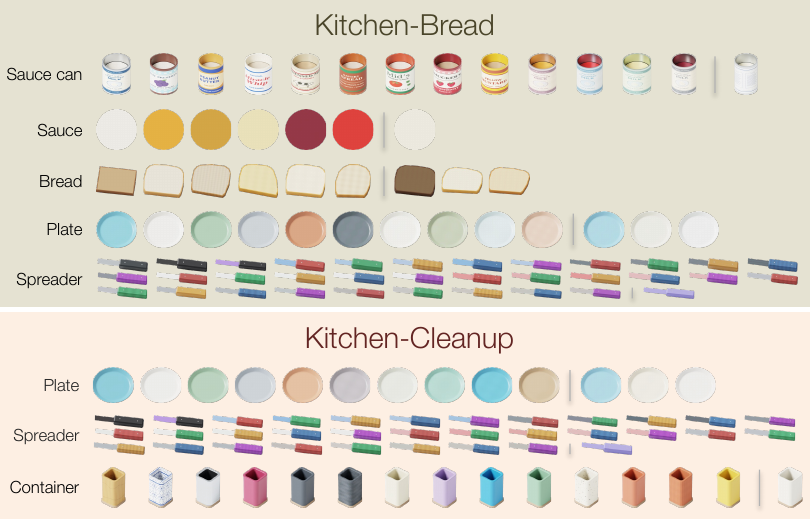}
    \caption{\small
    \textbf{Simulated objects of the tasks.}
    Each row shows the appearances of one object category in the
    \texttt{Kitchen-Bread} and \texttt{Kitchen-Cleanup} tasks on
    \textsc{G1-Sim-Dex3}. Appearances to the right of the divider in each row
    are used only in OOD trials, and the others are randomized in the
    demonstrations.
    }
    \label{fig:objects_sim}
\end{figure}

\paragraph{ID and OOD conditions.}
Both conditions are defined for each task--embodiment pair with respect to its
fine-tuning dataset, i.e., the demonstrations of the target task on the target
embodiment, which are excluded from pretraining
(Appendix~\ref{sec:app_algorithm}). \emph{In-distribution (ID)} trials use
settings drawn from the same distribution as the fine-tuning demonstrations, including the same objects and receptacles and the same range of initial object placements. \emph{Out-of-distribution (OOD)}
trials use configurations that are unseen in the fine-tuning dataset, such as
new object shapes, sizes, and colors, a different receptacle, sauce, or plate,
or new initial object placement, while the task goal and the stage scoring remain
unchanged. The ID and OOD configurations of each pair are given after the
description of each task below. Fig.~\ref{fig:objects_real} shows the real
objects of each task, and Fig.~\ref{fig:objects_sim} the object appearances in
simulation.

\paragraph{OOD evaluation tests cross-embodiment transfer.}
OOD configurations are absent from the target task–embodiment fine-tuning data. Some occur in the cross-embodiment pretraining mixture: in the
demonstrations of other embodiments, most notably the human demonstrations,
or in demonstrations of other tasks. A policy trained only on the
target-embodiment data, such as ACT or MED without pretraining, has therefore
never observed these configurations, whereas a pretrained policy may have
observed them while they were handled by a human or another robot. The OOD
evaluation is designed to measure exactly this case: \emph{whether experience
with a configuration on other embodiments, especially humans, transfers to a
target robot that has never encountered it}. This is the setting that
cross-embodiment pretraining targets, and OOD results should be read as a test
of cross-embodiment transfer rather than of generalization to configurations
that are absent from all training data.

\paragraph{\texttt{Scoop} (cat litter scooping).}
The scene contains a litter box filled with cat litter, a scoop resting on the
edge of the box, and a disposal bin next to it. The robot grasps the scoop,
moves it to the waste in the box, i.e., the target object, scoops and lifts
it, empties it into the disposal bin, and returns the scoop to the edge of the
box. On \textsc{xArm7-Dex4} and \textsc{Z1-Tool}, the box stands on a table.
On \textsc{G1-Dex5}, it is placed on the floor, so the humanoid lowers its
torso to reach it, and \textsc{LocoMan-Dex2} holds the scoop with the gripper
on its front leg. \textsc{Z1-Tool} carries a scoop that is rigidly attached to
the arm and therefore skips the grasp.
\emph{ID and OOD.} On all four embodiments, ID trials use a white disposal bin
within the position range seen in fine-tuning, and OOD trials replace it with a
green disposal bin or move it to new positions. On \textsc{G1-Dex5},
\textsc{xArm7-Dex4}, and \textsc{Z1-Tool}, whose workspaces are large, OOD
trials also place the waste at new positions in the litter box.

\paragraph{\texttt{Sort} (block sorting).}
Red and blue building blocks lie on the table in front of a red and a blue
bin, whose left--right order varies across trials. The robot picks up each
block and places it into the bin of the same color. Each block is scored
independently, and approaching or placing a block into the wrong bin earns only
partial credit (Table~\ref{tab:task_progress}). The bimanual
\textsc{G1-Dex5} and \textsc{R1Lite-Gripper} use both arms.
\emph{ID and OOD.} ID trials use three blocks of the shapes seen in the
fine-tuning data, and each reproduces the initial layout of a recorded
demonstration. On all three embodiments, OOD trials use blocks of new shapes,
e.g., a four-stud brick. On \textsc{R1Lite-Gripper}, OOD trials also change
the number of blocks, e.g., to four.

\paragraph{\texttt{Insert} (T-shaped block insertion).}
A T-shaped block lies on the table next to a fixed base with a T-shaped slot.
The robot approaches and grasps the block, reorients and aligns it with the
slot, and inserts it; a near-successful insertion earns partial credit.
\emph{ID and OOD.} ID trials use the block seen in the fine-tuning data, which
is light purple on \textsc{G1-Dex5} and white on \textsc{xArm7-Dex4}. On
\textsc{xArm7-Dex4}, the block is placed at nine predefined positions, one of
which is repeated four more times. OOD trials on \textsc{G1-Dex5} use a white
block, a dark purple block, or a larger brown block, and OOD trials on
\textsc{xArm7-Dex4} use the light purple block or the larger brown block. For each embodiment, the OOD evaluation uses blocks demonstrated by the other embodiment during cross-embodiment pretraining.

\paragraph{\texttt{Kitchen-Bread} (sauce spreading).}
This long-horizon task combines locomotion with manipulation. On
\textsc{G1-Dex5}, the robot opens the refrigerator, grasps the sauce
container, closes the refrigerator, walks to the counter, and places the
container next to a plate of bread. It then picks up the spreading tool, dips
it into the sauce, spreads the sauce onto the bread, and puts the tool back.
On \textsc{R1Lite-Gripper}, the robot picks up the sauce container from a
counter instead of a refrigerator, drives its wheeled base to the table with
the bread, places the container, and performs the same spreading steps.
\textsc{G1-Sim-Dex3} performs the \textsc{G1-Dex5} sequence in the simulated
kitchen (Appendix~\ref{app:simulation}).
\emph{ID and OOD.} On \textsc{G1-Dex5}, ID trials use white condensed milk and
brown bread on a green plate, and OOD trials use another kind of condensed
milk and a blue plate, with white or brown bread. On \textsc{R1Lite-Gripper},
ID trials use white condensed milk and brown bread, and OOD trials use a
different sauce, e.g., mayonnaise or strawberry jam, or white condensed milk
on white bread; the plate is the same in both conditions.
On \textsc{G1-Sim-Dex3}, ID trials sample object appearances and layouts from the demonstration distribution. OOD trials use unseen object and scene appearances, together with unseen plate--spreader layouts (Appendix~\ref{app:simulation}).

\paragraph{\texttt{Kitchen-Cleanup} (tool organization).}
This task is evaluated only on \textsc{G1-Sim-Dex3}. A spreader rests on a
plate on the counter next to a container, and the robot grasps the spreader
and places it into the container.
\emph{ID and OOD.} ID trials sample appearances and layouts from the demonstration distribution. OOD trials use object and scene appearances not present in those demonstrations (Appendix~\ref{app:simulation}).

\paragraph{Trials and scoring.}
Each real-world pair is evaluated over 20 trials (13 ID and 7 OOD), and each
simulated pair over 50 trials (32 ID and 18 OOD). All methods are evaluated
on the same trial configurations. We score each trial with a stage-based task
progress score that gives partial credit for intermediate stages. For a trial
with accumulated stage score $s$, success indicator $b\in\{0,1\}$, and
task-specific maximum score $S_{\max}$ including the success bonus, the
normalized task progress score is $P=(s+b)/S_{\max}$.
Table~\ref{tab:task_progress} lists the stages, points, and success criteria
of each task. ID and OOD scores are reported separately.

\begin{table*}[h]
    \centering
    \caption{
    \textbf{Task progress scoring used for evaluation.}
    A trial earns the points of every stage it completes and a one-point
    success bonus if it succeeds. $S_{\max}$ is the full stage score plus the
    bonus, and the normalized task progress score is $P=(s+b)/S_{\max}$
    (Appendix~\ref{app:evaluation_protocol}).
    }
    \label{tab:task_progress}

    \small
    \setlength{\tabcolsep}{5pt}
    \renewcommand{\arraystretch}{1.0}

    \begin{tabularx}{\textwidth}{
        @{}
        l
        >{\raggedright\arraybackslash}p{2.75cm}
        >{\raggedright\arraybackslash}X
        c
        @{}
    }
        \toprule
        \textbf{Task}
        & \textbf{Embodiment}
        & \textbf{Stages and points}
        & $\mathbf{S_{\max}}$ \\
        \midrule

        \texttt{Scoop}
        & \textsc{G1-Dex5}, \textsc{xArm7-Dex4}, \textsc{LocoMan-Dex2}
        & Grasp the scoop: 1;
        move the scoop to the target object: 1;
        scoop and lift the object: 2;
        drop the object into the bin: 1.
        & 5\,+\,1 \\

        & \textsc{Z1-Tool}
        & The same stages without the grasp, since the scoop is fixed to the
        arm: approach: 1; scoop and lift: 2; deposit: 1.
        & 4\,+\,1 \\
        \midrule

        \texttt{Sort}
        & \textsc{G1-Dex5}, \textsc{xArm7-Dex4}, \textsc{R1Lite-Gripper}
        & Each of the $n$ blocks is scored independently:
        pick up the block: 1;
        approach a bin: 3 for the bin of the block's color and 1 for the
        other bin;
        place the block into a bin: 3 for the bin of the block's color and 1
        for the other bin.
        All trials use $n=3$ blocks, except \textsc{R1Lite-Gripper} OOD
        trials with more blocks, e.g., $n=4$; each trial is normalized by its
        own $S_{\max}$.
        & $7n$\,+\,1 \\
        \midrule

        \texttt{Insert}
        & \textsc{G1-Dex5}, \textsc{xArm7-Dex4}
        & Approach the T-shaped block: 1;
        grasp the block: 2;
        insert the block into the slot: 2 for a full insertion, 1 for a
        near-successful insertion, and 0 otherwise.
        & 5\,+\,1 \\
        \midrule

        \texttt{Kitchen-Bread}
        & \textsc{R1Lite-Gripper}
        & Pick up the sauce container: 1;
        drive to the table: 1;
        place the container: 1;
        pick up the spreading tool: 1;
        dip the tool into the sauce: 1;
        spread the sauce onto the bread: 1;
        put the tool back: 1.
        & 7\,+\,1 \\

        & \textsc{G1-Dex5}, \textsc{G1-Sim-Dex3}
        & Open the refrigerator: 1;
        grasp the sauce container: 1;
        close the refrigerator: 1;
        walk to the counter: 1;
        place the container: 1;
        pick up the spreading tool: 1;
        dip the tool into the sauce: 1;
        spread the sauce onto the bread: 1;
        put the tool back: 1.
        & 9\,+\,1 \\
        \midrule

        \texttt{Kitchen-Cleanup}
        & \textsc{G1-Sim-Dex3}
        & Grasp the spreader: 1;
        lift the spreader: 1;
        insert the spreader into the container: 1;
        release the spreader so that it is supported inside the container: 1.
        & 4\,+\,1 \\

        \bottomrule
    \end{tabularx}

    \vspace{3pt}
    \begin{minipage}{\textwidth}
    \footnotesize
    On the physical robots, a trial succeeds, and earns the bonus, exactly when
    it completes every stage, so $P=1$ if and only if the trial succeeds. In
    simulation, \texttt{Kitchen-Bread} succeeds when the sauce added to the
    bread covers at least 5\% of it and the spreader ends released and stably
    supported, and \texttt{Kitchen-Cleanup} succeeds when the spreader is
    released and supported inside the container
    (Appendix~\ref{app:simulation}). A successful simulated
    \texttt{Kitchen-Bread} trial therefore need not complete every stage; for
    example, it may leave the refrigerator open.
    \end{minipage}
\end{table*}

\clearpage
\subsection{Success Rates}
\label{app:success_rate}

\begin{figure}[!htb]
    \centering
    \includegraphics[width=0.95\textwidth]{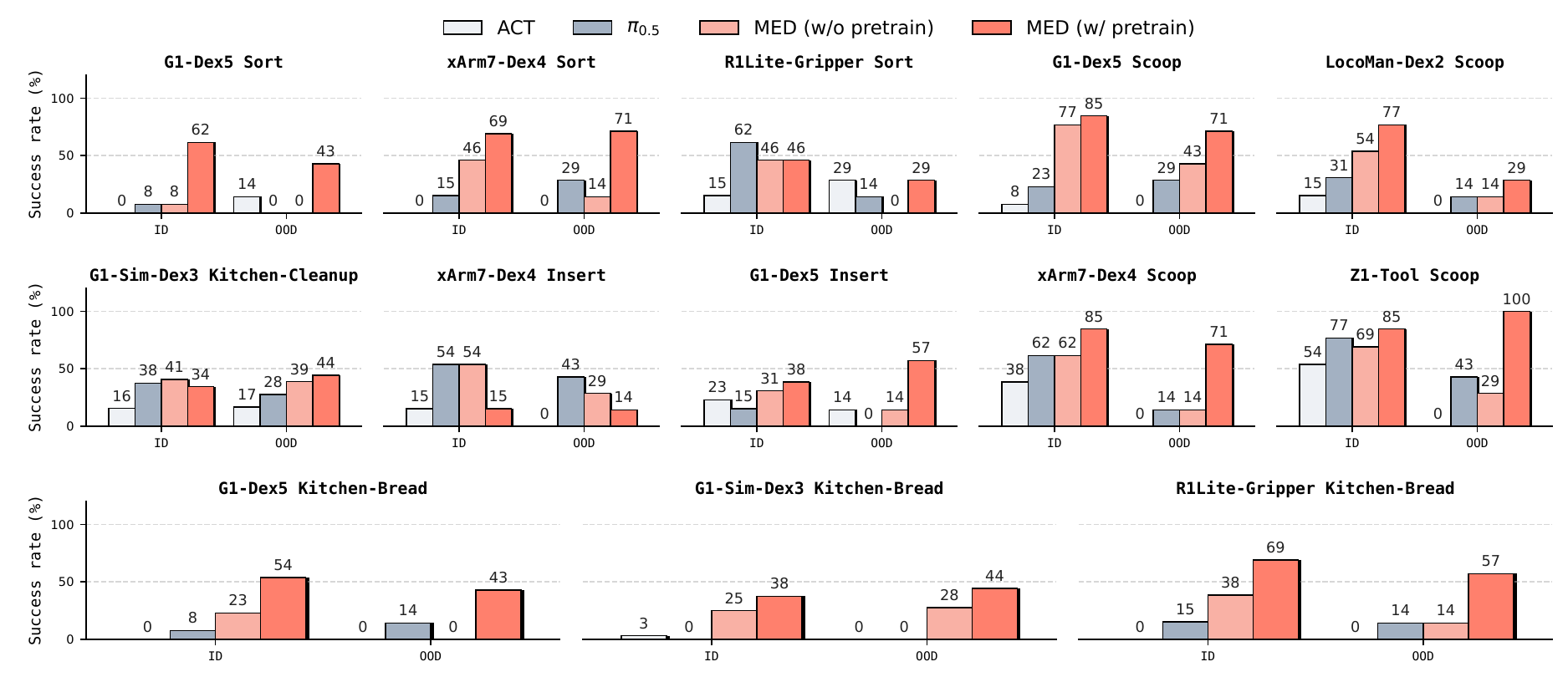}
    \caption{\small
    \textbf{Success rates.}
    Success rate of ACT, $\pi_{0.5}$, MED without pretraining, and MED with
    cross-embodiment pretraining on each task--embodiment pair under ID and OOD
    conditions, evaluated on the same trials as Fig.~\ref{fig:main_results}.
    }
    \label{fig:success_rate}
\end{figure}

Sec.~\ref{sec:experiments} reports the normalized task progress score $P$.
Fig.~\ref{fig:success_rate} reports the success rate on the same trials, i.e.,
the fraction of trials with $b=1$ (Appendix~\ref{app:evaluation_protocol}),
for each method and task--embodiment pair under ID and OOD conditions.

Success rates give the same overall ranking as task progress. Averaged over the
13 pairs, cross-embodiment pretraining raises the success rate of MED from 35\%
to 56\% and improves 11 of 13 pairs, while $\pi_{0.5}$ and ACT reach 27\% and
11\%. The gain is largest out of distribution: there, MED without pretraining
succeeds no more often than $\pi_{0.5}$ (18\% vs.\ 19\%), and pretraining raises
it to 52\%, compared with a rise from 44\% to 58\% in distribution. Embodiment
dreaming likewise raises the success rate over MXT-FG in all seven settings of
Fig.~\ref{fig:ablation_dreaming}, from 39\% to 62\% on average (from 46\% to
63\% in distribution and from 27\% to 60\% out of distribution), most on
\texttt{G1-Dex5 Scoop} (from 3 to 16 of 20 trials).

Tables~\ref{tab:sr_composition}, \ref{tab:sr_dreaming}, and~\ref{tab:sr_coarse} report the success rates of the pretraining data composition, embodiment dreaming, and coarse ablations (Figs.~\ref{fig:ablation_composition}, \ref{fig:ablation_dreaming}, and~\ref{fig:ablation_coarse}).

\begin{table}[!htb]
    \centering
    \caption{
    \textbf{Success rates of the pretraining data composition ablation.}
    Success rate on the trials of Fig.~\ref{fig:ablation_composition} as pretraining data is added: human, other robots, then the target robot's other tasks.
    The real robots use 13 ID and 7 OOD trials per mixture, and \textsc{G1-Sim-Dex3} uses 32 ID and 18 OOD trials.
    The last row averages the six pairs.
    Bold marks the best mixture in each row, separately for ID and OOD.
    }
    \label{tab:sr_composition}
    \fontsize{8pt}{9.6pt}\selectfont
    \setlength{\tabcolsep}{4pt}
    \renewcommand{\arraystretch}{1.15}
    \begin{tabular}{@{}llcccccccc@{}}
        \toprule
        & & \multicolumn{4}{c}{\textbf{ID}} & \multicolumn{4}{c}{\textbf{OOD}} \\
        \cmidrule(lr){3-6} \cmidrule(l){7-10}
        \textbf{Embodiment} & \textbf{Task}
        & \textbf{None} & \textbf{Human} & \shortstack{\textbf{+Other}\\\textbf{Robots}} & \shortstack{\textbf{+Target}\\\textbf{Robot}}
        & \textbf{None} & \textbf{Human} & \shortstack{\textbf{+Other}\\\textbf{Robots}} & \shortstack{\textbf{+Target}\\\textbf{Robot}} \\
        \midrule
        \textsc{G1-Dex5} & \texttt{Sort} & \phantom{0}8\% & 38\% & 54\% & \textbf{62\%} & \phantom{0}0\% & \phantom{0}0\% & \textbf{43\%} & \textbf{43\%} \\
        & \texttt{Insert} & 31\% & 31\% & \textbf{38\%} & \textbf{38\%} & 14\% & 29\% & 43\% & \textbf{57\%} \\
        \addlinespace[2pt]
        \textsc{xArm7-Dex4} & \texttt{Sort} & 46\% & 46\% & 46\% & \textbf{69\%} & 14\% & 14\% & 43\% & \textbf{71\%} \\
        & \texttt{Insert} & 54\% & \phantom{0}8\% & \textbf{62\%} & 15\% & 29\% & 29\% & \textbf{43\%} & 14\% \\
        \addlinespace[2pt]
        \textsc{R1Lite-Gripper} & \texttt{Sort} & \textbf{46\%} & 15\% & 15\% & \textbf{46\%} & \phantom{0}0\% & \textbf{29\%} & 14\% & \textbf{29\%} \\
        \addlinespace[2pt]
        \textsc{G1-Sim-Dex3} & \texttt{Kitchen-Bread} & 25\% & 19\% & 31\% & \textbf{38\%} & 28\% & 39\% & 17\% & \textbf{44\%} \\
        \midrule
        \multicolumn{2}{@{}l}{Average} & 35\% & 26\% & 41\% & \textbf{45\%} & 14\% & 23\% & 34\% & \textbf{43\%} \\
        \bottomrule
    \end{tabular}
\end{table}

\begin{table}[!htb]
    \centering
    \caption{
    \textbf{Success rates of the embodiment dreaming ablation.}
    Success rate of MXT-FG (w/o dreaming) and MED (w/ dreaming), both pretrained and fine-tuned, on the trials of Fig.~\ref{fig:ablation_dreaming}.
    The real robots use 13 ID and 7 OOD trials per method, and \textsc{G1-Sim-Dex3} uses 32 ID and 18 OOD trials.
    Overall counts both ID and OOD trials, as in Fig.~\ref{fig:ablation_dreaming}.
    The last row averages the seven pairs.
    Bold marks the better method in each row, separately for ID, OOD, and overall.
    }
    \label{tab:sr_dreaming}
    \fontsize{8pt}{9.6pt}\selectfont
    \setlength{\tabcolsep}{4pt}
    \renewcommand{\arraystretch}{1.15}
    \begin{tabular}{@{}llcccccc@{}}
        \toprule
        & & \multicolumn{2}{c}{\textbf{ID}} & \multicolumn{2}{c}{\textbf{OOD}} & \multicolumn{2}{c}{\textbf{Overall}} \\
        \cmidrule(lr){3-4} \cmidrule(lr){5-6} \cmidrule(l){7-8}
        \textbf{Embodiment} & \textbf{Task}
        & \textbf{w/o} & \textbf{w/} & \textbf{w/o} & \textbf{w/} & \textbf{w/o} & \textbf{w/} \\
        \midrule
        \textsc{G1-Dex5} & \texttt{Scoop} & 15\% & \textbf{85\%} & 14\% & \textbf{\phantom{0}71\%} & 15\% & \textbf{80\%} \\
        & \texttt{Insert} & 15\% & \textbf{38\%} & 29\% & \textbf{\phantom{0}57\%} & 20\% & \textbf{45\%} \\
        \addlinespace[2pt]
        \textsc{xArm7-Dex4} & \texttt{Scoop} & \textbf{85\%} & \textbf{85\%} & 29\% & \textbf{\phantom{0}71\%} & 65\% & \textbf{80\%} \\
        \addlinespace[2pt]
        \textsc{G1-Sim-Dex3} & \texttt{Kitchen-Bread} & 19\% & \textbf{38\%} & 11\% & \textbf{\phantom{0}44\%} & 16\% & \textbf{40\%} \\
        & \texttt{Kitchen-Cleanup} & 31\% & \textbf{34\%} & 22\% & \textbf{\phantom{0}44\%} & 28\% & \textbf{38\%} \\
        \addlinespace[2pt]
        \textsc{LocoMan-Dex2} & \texttt{Scoop} & 69\% & \textbf{77\%} & \phantom{0}0\% & \textbf{\phantom{0}29\%} & 45\% & \textbf{60\%} \\
        \addlinespace[2pt]
        \textsc{Z1-Tool} & \texttt{Scoop} & \textbf{85\%} & \textbf{85\%} & 86\% & \textbf{100\%} & 85\% & \textbf{90\%} \\
        \midrule
        \multicolumn{2}{@{}l}{Average} & 46\% & \textbf{63\%} & 27\% & \textbf{\phantom{0}60\%} & 39\% & \textbf{62\%} \\
        \bottomrule
    \end{tabular}
\end{table}

\begin{table}[!htb]
    \centering
    \caption{
    \textbf{Success rates of the coarse ablation.}
    Success rate of MED and its coarse variant without (w/o) and with (w/) cross-embodiment pretraining, on the trials of Fig.~\ref{fig:ablation_coarse}.
    The real robots use 13 ID and 7 OOD trials per method, and \textsc{G1-Sim-Dex3} uses 32 ID and 18 OOD trials.
    The last row averages the four pairs.
    Bold marks the best result in each row, separately for ID and OOD.
    }
    \label{tab:sr_coarse}
    \fontsize{8pt}{9.6pt}\selectfont
    \setlength{\tabcolsep}{4pt}
    \renewcommand{\arraystretch}{1.15}
    \begin{tabular}{@{}llcccccccc@{}}
        \toprule
        & & \multicolumn{4}{c}{\textbf{ID}} & \multicolumn{4}{c}{\textbf{OOD}} \\
        \cmidrule(lr){3-6} \cmidrule(l){7-10}
        & & \multicolumn{2}{c}{\textbf{MED}} & \multicolumn{2}{c}{\textbf{MED (coarse)}} & \multicolumn{2}{c}{\textbf{MED}} & \multicolumn{2}{c}{\textbf{MED (coarse)}} \\
        \cmidrule(lr){3-4} \cmidrule(lr){5-6} \cmidrule(lr){7-8} \cmidrule(l){9-10}
        \textbf{Embodiment} & \textbf{Task}
        & \textbf{w/o} & \textbf{w/} & \textbf{w/o} & \textbf{w/}
        & \textbf{w/o} & \textbf{w/} & \textbf{w/o} & \textbf{w/} \\
        \midrule
        \textsc{Z1-Tool} & \texttt{Scoop} & 69\% & \textbf{85\%} & 62\% & \textbf{85\%} & 29\% & \textbf{100\%} & 14\% & 86\% \\
        \addlinespace[2pt]
        \textsc{xArm7-Dex4} & \texttt{Scoop} & 62\% & \textbf{85\%} & 15\% & \phantom{0}0\% & 14\% & \textbf{\phantom{0}71\%} & 14\% & \phantom{0}0\% \\
        \addlinespace[2pt]
        \textsc{G1-Sim-Dex3} & \texttt{Kitchen-Bread} & 25\% & \textbf{38\%} & 13\% & \phantom{0}0\% & 28\% & \textbf{\phantom{0}44\%} & 17\% & \phantom{0}0\% \\
        & \texttt{Kitchen-Cleanup} & \textbf{41\%} & 34\% & 34\% & \phantom{0}0\% & 39\% & \textbf{\phantom{0}44\%} & 22\% & \phantom{0}0\% \\
        \midrule
        \multicolumn{2}{@{}l}{Average} & 49\% & \textbf{60\%} & 31\% & 21\% & 27\% & \textbf{\phantom{0}65\%} & 17\% & 21\% \\
        \bottomrule
    \end{tabular}
\end{table}

\subsection{Simulation Details}
\label{app:simulation}
\paragraph{Simulation setup.}
We simulate \texttt{G1-Sim-Dex3}, a Unitree G1 humanoid with three-finger Dex3 hands, in a kitchen environment built in Isaac Sim and Isaac Lab. In \texttt{Kitchen-Bread}, the robot retrieves a sauce can from the refrigerator, carries it to the counter, and spreads its contents onto bread; the shorter \texttt{Kitchen-Cleanup} task requires picking up the spreader and placing it in a container. Objects interact through rigid-body contact, while sauce transfer is rendered visually without fluid simulation. Demonstrations are collected by teleoperation with PICO controllers, and the egocentric head-camera and wrist-camera images are rendered offline with RTX. Physics runs at $200\,\mathrm{Hz}$ and the locomotion controller at $50\,\mathrm{Hz}$.

\paragraph{Randomization.}
At reset, we randomize the robot and object positions, object sizes, aisle width, and counter height, keep the spreader supported by the plate, and check each sampled layout for clearance. We also vary the bread shape and the positions of background objects and furniture. During offline rendering, we randomize object and room materials, lighting, camera poses and fields of view, and image properties such as color, noise, blur, and compression. These visual changes leave the recorded motions and action labels unchanged. Each rendered episode uses a fixed visual configuration, with noise varying across frames, and the dataset also retains variants close to the original appearance. Fig.~\ref{fig:objects_sim} shows the object appearances, including those used only in OOD trials.

\paragraph{Evaluation.}
Each task is evaluated on 50 fixed trials: 32 ID and 18 OOD. OOD trials use object and scene appearances excluded from the demonstrations; \texttt{Kitchen-Bread} OOD trials also include plate–spreader arrangements outside the demonstrated range. Trial configurations were refined through pilot evaluations and then fixed for all methods. \texttt{Kitchen-Bread} and \texttt{Kitchen-Cleanup} trials last 90 and 45 simulated seconds, respectively. All methods execute actions at $30\,\mathrm{Hz}$, predict 60-action chunks, and replan every 45 action steps; $\pi_{0.5}$ uses RTC (Appendix~\ref{app:baselines}). \texttt{Kitchen-Bread} success requires spreading sauce onto the bread and, at the end of the trial, a released spreader settled on a support. \texttt{Kitchen-Cleanup} success requires grasping the spreader and placing it inside the container, released and supported, at any point during the trial. We score nine prerequisite-dependent stages for \texttt{Kitchen-Bread} and four for \texttt{Kitchen-Cleanup}, plus one point for success, giving maximum scores of 10 and 5; normalizing by these maxima gives the task progress score reported in Sec.~\ref{sec:experiments}.

\end{document}